\documentclass[utf8]{FrontiersinHarvard} % for articles in journals using the Harvard Referencing Style (Author-Date), for Frontiers Reference Styles by Journal: https://zendesk.frontiersin.org/hc/en-us/articles/360017860337-Frontiers-Reference-Styles-by-Journal
\usepackage{caption}
\usepackage{subcaption}
\usepackage{tabularx}
\usepackage{footnote}
\usepackage{url,hyperref,lineno}
\usepackage[onehalfspacing]{setspace}
\usepackage{graphicx}
\usepackage{amsmath}
\usepackage{natbib}
\usepackage{soul}
\usepackage{xcolor}

\setcitestyle{authoryear}
\def\keyFont{\fontsize{8}{11}\helveticabold }
\def\firstAuthorLast{Zhang} %use et al only if is more than 1 author
\def\Authors{Haiyun Zhang\,$^{1,*}$, Gabrielle Naquila\,$^{1}$, Jung Hyun Bae\,$^{1}$,  Zonghuan Wu\,$^{2}$,  Ashwin Hingwe\,$^{1}$, and Ashish Deshpande\,$^{1}$}

\def\Address{$^{1}$Mechanical Engineering, The University of Texas at Austin, Austin, TX, USA , \\$^{2}$ School of Integrated Circuits, Tsinghua University, 100190, Beijing, China}
\def\corrAuthor{Haiyun Zhang, Ashish Deshpande}

\def\corrEmail{haiyunzhang@utexas.edu, ashish@austin.utexas.edu}

\begin{document}
\onecolumn
\firstpage{1}

\title{Novel bio-inspired soft actuators for upper-limb exoskeletons: design, fabrication and feasibility study}

\author[\firstAuthorLast ]{\Authors} % This field will be automatically populated
\address{} % This field will be automatically populated
\correspondance{} % This field will be automatically populated

\extraAuth{} % If there are more than 1 corresponding author, comment this line and uncomment the next one.
%\extraAuth{Corresponding Author2 \\ Laboratory X2, Institute X2, Department X2, Organization X2, Street X2, City X2 , State XX2 (only USA, Canada and Australia), Zip Code2, X2 Country X2, email2@uni2.edu}

\maketitle

\thanks{Haiyun Zhang$^{*}$ (Corresponding Author) contributed to the conceptualization, design, fabrication, experiments, and overall coordination of the project, as well as the preparation of the manuscript. Email: haiyunzhang@utexas.edu}\newline
\thanks{Gabrielle Naquila was responsible for academic writing, graphing, and data analysis. Email: gabrielle.naquila@utexas.edu}\newline
\thanks{Jung Hyun Bae worked on the detailed mechanical design and fabrication processes of the actuators and the graphing of 3D model graphs. Email: jhbae131@utexas.edu}\newline
\thanks{Zonghuan Wu provides technical support on sensors and embedded systems. Email: wu-zh20@mails.tsinghua.edu.cn}\newline
\thanks{Ashwin Hingwe designed wearable mechanical settings to mount the actuators on experimental platforms and human subjects. Email: ashwinh@utexas.edu}\newline
\thanks{Ashish Deshpande, the leader of ReNeu Lab, supervised the project, provided critical feedback, and helped shape the research direction and analysis. Email: ashish@austin.utexas.edu}

%%% Leave the Abstract empty if your article does not require one, please see the Summary Table for full details.
% \section{}
\begin{abstract}
Soft robots have been increasingly utilized as sophisticated tools in physical rehabilitation, particularly for assisting patients with neuromotor impairments. However, many soft robotics for rehabilitation applications are characterized by limitations such as slow response times, restricted range of motion, and low output force. There are also limited studies on the precise position and force control of wearable soft actuators. Furthermore, not many studies articulate how bellow-structured actuator designs quantitatively contribute to the robots' capability. This study introduces a paradigm of upper limb soft actuator design. This paradigm comprises two actuators: the Lobster-Inspired Silicone Pneumatic Robot (LISPER) for the elbow and the Scallop-Shaped Pneumatic Robot (SCASPER) for the shoulder. LISPER is characterized by higher bandwidth, increased output force/torque, and high linearity. SCASPER is characterized by high output force/torque and simplified fabrication processes. Comprehensive analytical models that describe the relationship between pressure, bending angles, and output force for both actuators were presented so the geometric configuration of the actuators can be set to modify the range of motion and output forces. The preliminary test on a dummy arm is conducted to test the capability of the actuators.

\tiny
 \keyFont{ \section{Keywords:} Index Terms—Pneumatic soft actuators, bio-inspired design, analytical modeling, wearable devices}
\end{abstract}

\section{Introduction}
Soft robotics, in contrast to traditional rigid robots, are often considered safer for human-robot interaction due to their inherent compliant properties. This characteristic is particularly crucial for rehabilitation applications \citep{o2017soft,fang2020novel}. Soft robots also exhibit a high power-to-weight ratio \citep{o2022unfolding} and are typically lighter than rigid exoskeletons, which may enable soft actuator-based exoskeletons to operate in a more power-efficient manner. 

This lightweight nature, combined with the inherent flexibility of soft materials, enhances the wearability and comfort of wearable exoskeleton devices. The improved wearability not only reduces user fatigue but also increases mobility, making these devices easier to transport and wear for extended periods. Consequently, soft wearable robots are well-suited for use in daily living environments, supporting rehabilitation and assistance in a more practical and user-friendly way. Additionally, developing soft robots is generally more cost-effective compared to manufacturing most rigid exoskeletons.

In rehabilitation training, there are two categories of soft robots commonly implemented tendon-driven and pneumatic-driven. Tendon-driven exosuits use strings and ropes as transmission and soft materials as interfaces between human tissue and the end effect point of the transmission. These systems can achieve a large range of motion, precise force, and position control, and generate large output force/ torques~\cite{yun2017emg,yun2020improvement,yun2017maestro}. However, several issues originate from tendon-driven soft exoskeletons. The strings connecting the tendon-driven actuators to the motors are prone to stretching, which requires them to be housed in isolated spaces to prevent unintended contact with the surrounding environment~\cite{yun2017maestro}. Furthermore, the movement of a person's limbs can also shift the position of the strings' ends, leading to unwanted slack or tension in the strings. Tendon-driven joints also need at least two strands of strings to realize bidirectional motion with only one pipe given the elastic (or hyper-elastic) property of the actuator body~\citep{gerez2019development}. 

Pneumatically driven soft actuators are commonly used in human-involved experiments due to their intrinsic compliance and easy setting-up process. \cite{heung2019robotic} developed a silicone rubber-based beam-shaped wearable glove to assist with post-stroke rehabilitation training. Similarly, \cite{park2022fabric} introduced a fabric-based pneumatic exoskeleton for elbow-assistive motion. Natividad et al. introduced an inflatable, fabric beam shoulder actuator emphasizing lightweight and simple structural design \cite{7523758}. However, this design result is bulky and has limited output force. An extension from Natividad et al. introduces another design aiming to cover both the shoulder and elbow with a single mechanism \cite{9291462}. This design occupies substantial volume and requires numerous pneumatic actuator modules, making fabrication complex and cumbersome. There are other previous pneumatically-driven soft robotic systems for upper-limb assistive training. In general, these actuator designs, despite being suitable intrinsically for rehabilitation training, show limitations in assistive motion tasks including: 

\textbf{(1) Insufficient moment of force to effectively support human limb motion.} The typical range of force generated by silicone rubber-based soft robots is from 0 - 10 N ~\citep{huang2020variable}, less than the required force to push the human’s upper limb.

\textbf{(2) The fabrication process for soft robotics is time-consuming and requires complicated tools.} Many fabrications, especially the fabrication of inextensible fabric-based devices, require laser cutting and heat-sealing machines~\citep{nguyen2019fabric,9291462}.

\textbf{(3) Establishing an accurate analytical model for soft robots is difficult} due to the highly nonlinear property of the hyper-elastic material and the challenge of describing their irregular shape in analytical forms.  \citep{muller2020one,huang2020variable,7523758}.

\begin{figure}[htp]
    \centering
    \includegraphics[width=3.2in]{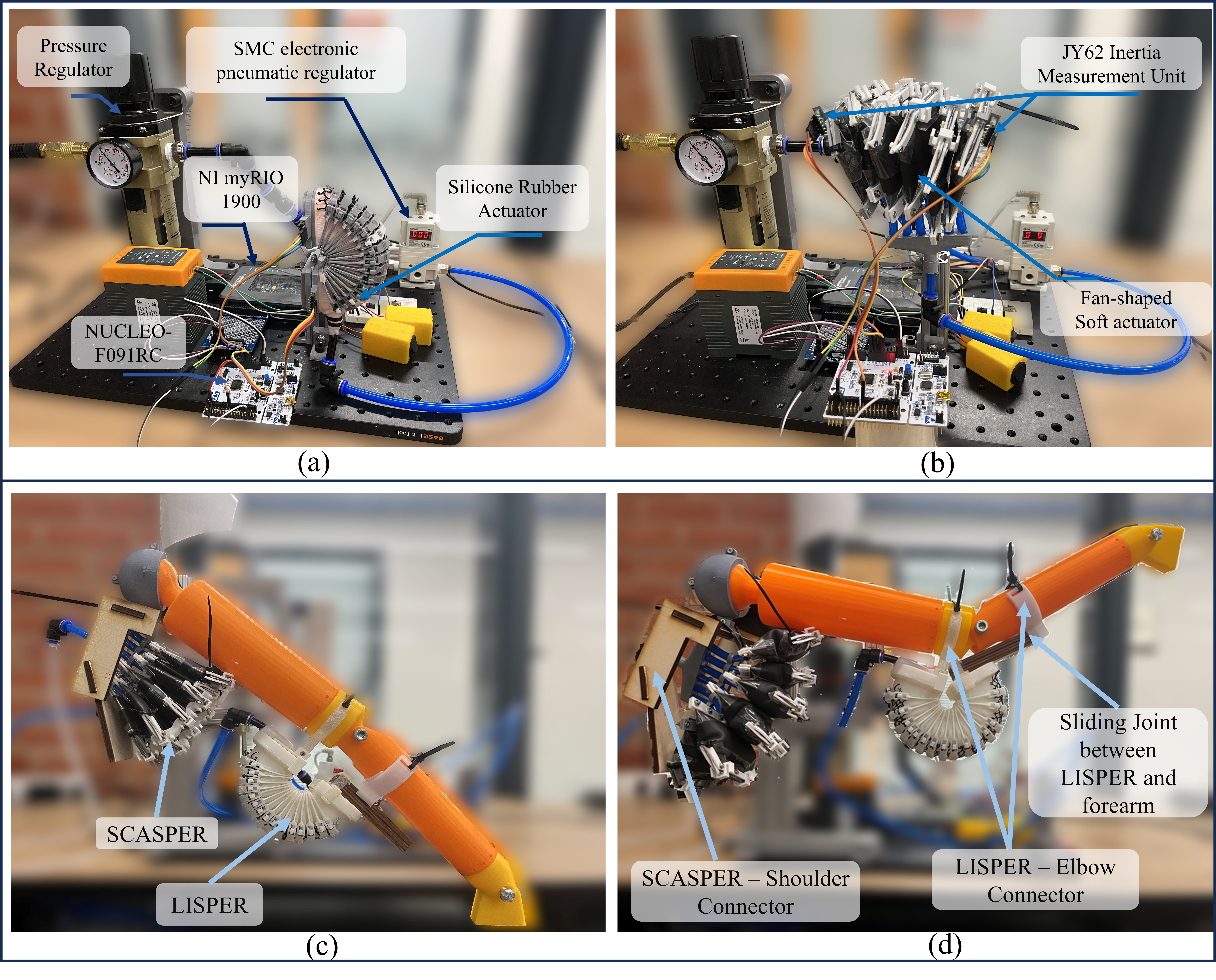}
    \caption{(a) Experimental Platform of LISPER for Unloading Bending Test (b) Experimental Platform of SCASPER for Unloading Bending Test (c) 2-DOF dummy arm before inflation. (d) 2-DOF dummy arm after inflation.}
\label{fig_1}
\end{figure}

In this study, we introduce a design paradigm of soft pneumatic actuators for upper-limb assistance comprised of two soft actuator designs  \textbf{Lobster-Inspired Silicone Pneumatic Robot (LISPER) } for below shown in Fig. \ref{fig_1} (a) \textbf{SCAllop-Shaped Pneumatic Robot (SCASPER)}  Fig. \ref{fig_1} (b) along with their analytical modeling to tackle the issues mentioned above. Lobster-Inspired Silicone Pneumatic Robot (LISPER) is inspired by the rigid crust of the lobster and from which we included the c-shape constraint to restrict the elongation of the outer rings of LISPER as shown in Fig.\ref{fig_1}. (a). Similarly, the Scallop-Shaped Pneumatic Robot (SCASPER) is inspired by the opening motion of scallops. When the adductor muscle of the SCASPER inflates, then the two fans of the scallop open (Fig.\ref{fig_1}. (b)). The model not only describes the relationship between pressure, bending angles, and output force but also explains how to change the range of motion and output forces numerically by modifying the geometry of the bellows.  The feasibility of the device is tested on a dummy arm by applying position and gravity compensation controls. 

In specific, this research makes contributions in three aspects:

\textbf{(1) In mechanical design}, we propose two types of pneumatic-driven soft actuators, Scallop-Shaped Pneumatic Robot (SCASPER) and Lobster-Inspired Silicone Pneumatic Robot (LISPER) for the shoulder and elbow, respectively. Both actuators outperform previous works in dynamic properties, including range of motion, maximum force output, linearity between pressure input and position output, low hysteresis, and high bandwidth, as mentioned above and in Table. \ref{table:tab2}. 

\textbf{(2) In analytical modeling}, we characterize the contribution of unfolding bellow structure by analytical modeling. This model can be used for parameter-driven mechanical design as it demonstrates how the triangular bellows design significantly enhances performance, providing additional forces and increased bending angles in response to pressure input. 

\textbf{(3) In the fabrication method}, we introduce a concise fabrication technique to make inextensible soft actuators (SCASPER) that do not require fabrication machines and are time-efficient. Besides, we elaborate on the fabrication techniques to avoid failures caused by stress concentration and air leakage, which are common in large soft actuators. 

Among the two actuators,  LISPER is designed with (1) precise bending motion without a sophisticated controller, (2) high linearity between bending angle and inflation pressure input, (3) higher bandwidth and faster response rate, and (4) the ability to generate larger force and moment of force than conventional pneumatic soft robots. 

However, the output force from the LISPER is insufficient to support motions of shoulder. The other actuator, SCASPER, is designed to solve this issue. Its features include (1) the ability to generate large output force/torques to support heavy human arms, (2) adjustable linearity and quicker deflation speed than conventional soft robots made of inextensible layers, and (3) low difficulty of fabrication and requiring only commonly used materials (e.g., polyethylene heat-shrinking tubes).

The research community could benefit from this work in two aspects (1) By directly following the design and fabrication process, rehabilitation studies could use the design for clinical research. (2) Upgrade upper-limb exoskeleton can be made following the design paradigm including c-shape constraints,  tunable elasticity from rubber strips,  two-actuator frame for the two joints, and parameter-driven geometric design.

\begin{figure*}[htbp]
\centering
\includegraphics[width=\textwidth]{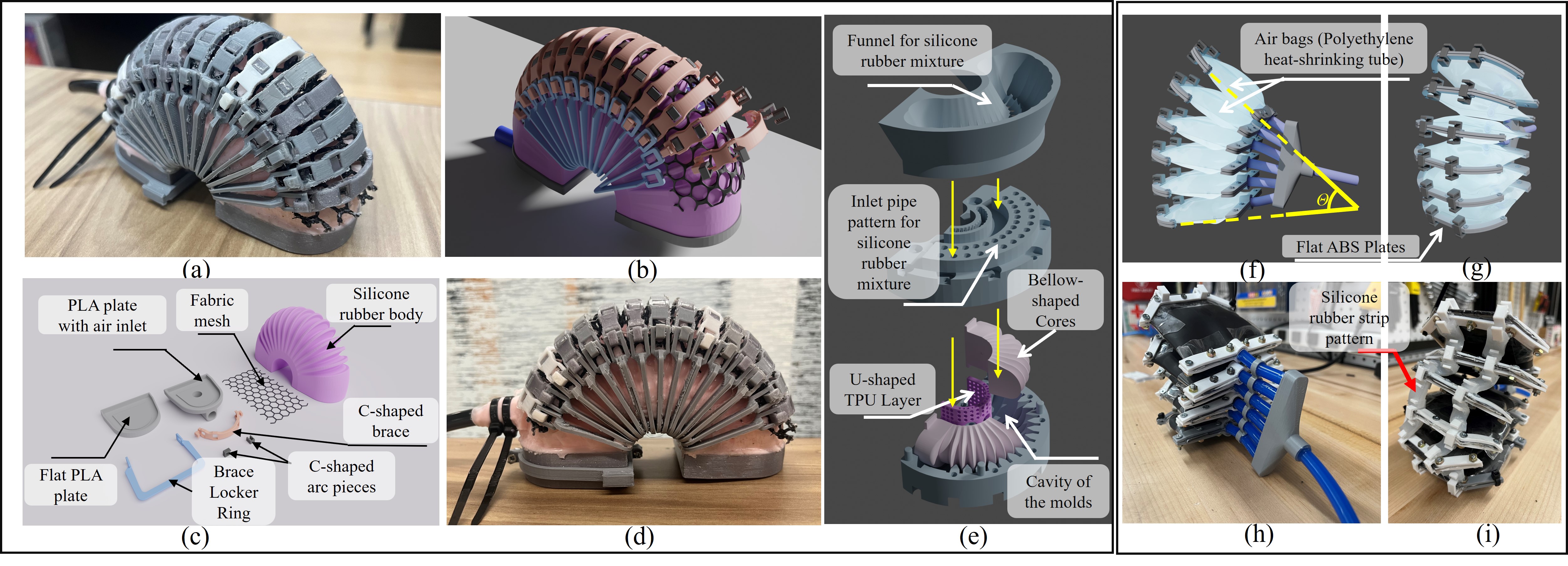}
\caption{(a) and (b) Actual and CAD designs of LISPER. (c) Demonstration of components of unassembled LISPER and (d) The side view of LISPER. (e) Exploded view of molding and casting procedure of LISPER. (f) and (g) CAD demonstration of SCASPER, $\theta$  is the extension angle. (h) and (i) are front and rear views of actual SCASPER}
\label{fig_2}
\end{figure*}

The following sections are as follows:

Section II elaborates on the mechanical design and fabrication of LISPER and SCASPER. This section also explains the simulation of two actuators with the finite element analysis method (FEA/FEM) and the comparison with experimental results. Section III introduces the analytical modeling of LISPER and SCASPER, and discusses the impact of the bellow structure and its quantitative contribution to the range of motion and output force. Section IV introduces how the experiments are conducted and compares the results with simulations. Section V discusses the details and limitations of the current work and explains prospective work. Finally, Section VI summarizes the conclusion.

\section{MECHANICAL DESIGN AND FABRICATION}
\subsection{Design Considerations}

The main purpose of the mechatronic design is to provide dynamic gravitational compensation for human subjects along upper-limb rehabilitation training procedures. Therefore, the metrics of the wearable design include (1) Large range of motion, (2) Significant assistive force, (3) High Response rate (4) Comfortable wearability.

To alleviate the difficulty of the fabrication process, the SCASPER is designed to be manufactured without machine tools. The main structure of the SCASPER shall be made of material low-cost and easy to obtain. The main configuration of LISPER and SCASPER are listed in Table. \ref{config1} and \ref{config2}. The size of the LISPER and SCASPER is defined by measuring the width of an adult male's forearm and rear arm, which are around 70 mm and 100 mm respectively.  The actuators are designed to be around 10-20\% shorter than these widths. This dimension setting aims to avoid bulkiness while providing sufficient output force/torque.

\begin{table*}[htbp]
\centering
\caption{Main Configuration of LISPER}
\begin{tabular}{|c|c|}
\hline
\textbf{Parameter}                         & \textbf{Value}            \\ \hline
Dimensions (mm)                            & 132 x 55 x 92             \\ \hline
Weight (g)                                 & 259                       \\ \hline
Range of Motion (Deg)                      & 112.2                     \\ \hline
Maximum Output                             & 12.5N/ 0.6 Nm             \\ \hline
Maximum Air Pressure (kPa)                 & 100                       \\\hline
\end{tabular}
\label{config1}
\end{table*}

\begin{table*}[htbp]
\centering
\caption{Main Configuration of SCASPER}
\begin{tabular}{|c|c|}
\hline
\textbf{Parameter}                         & \textbf{Value}            \\ \hline
Dimensions (mm)                            & 122 x 91 x 132            \\ \hline
Weight (g)                                 & 183                       \\ \hline
Range of Motion (Deg)                      & 122.5                     \\ \hline
Maximum Output                             & 49.5N/ 5.5Nm              \\ \hline
Maximum Air Pressure (kPa)                 & 150                       \\\hline
\end{tabular}
\label{config2}
\end{table*}

\subsection{Mechanical Design: Lobster-Inspired Silicone Pneumatic Robot (LISPER)}
LISPER (Fig. \ref{fig_2}(a)-(d)), drawing design inspiration from the morphology of lobster fins, is designed for motor joints like elbows and wrists that necessitate low output force/torque for operation, particularly in scenarios where gravitational force compensation is minimal. In a coordinated system, LISPER has the potential to collaborate with SCASPER to facilitate precise horizontal rotational movements, with SCASPER primarily addressing the substantial components of gravitational force. With these in mind, the design objectives established for LISPER included (1) achieving high linearity between applied pressure and bending angle in unloaded conditions and (2) ensuring a high response rate.

To achieve the design objectives, an arc-shaped body and bellows with sharp folds on one side of the external surface and the internal surface of the air chamber were incorporated into LISPER. These features serve several purposes: The arc-shaped design combined with bellow-shaped folds enables pure bending motion without undesired elongation, commonly occurring along the actuator's length. The sharp bellows tend to unfold under uniform pressure applied to the chamber's internally folded surface.
Bellow-shaped folds on the internal surface reduce the air chamber's volume, decreasing inflation and deflation times and reducing response time. The unfolding process of the bellow-shaped surface also accelerates bending speed. A minor benefit of an arc-shaped design is that it shifts the contact area between human tissue and the soft robot from the bottom to the side of the actuator, avoiding uncontrollable friction and the contact area between the surfaces. This change helps reduce sliding between surfaces, minimizing energy dissipation. 

In addition to the bellow and arc shape design, rigid-soft coupling was employed to constrain undesired radial extension and 'inverse folding' of the bellow-shaped folds. Brace locker and C-shape braces were added to the external edges of each fold, and a fabric mesh with hexagon patterns was placed between the brace and each fold's tips  (Fig. \ref{fig_2}(b)). When stretched in one direction, the hexagon fabric mesh shrinks in the perpendicular direction. This property achieves two goals: 1) providing an upper folding boundary to prevent inverse folding and 2) further decreasing undesired radial extension. A U-shaped 3D-printed TPU layer was added around the actuator's bottom to constrain elongation in that area and define the bending's neutral layer (Fig. \ref{fig_2}(e)). The actuator's two 'feet' were sealed with flat PLA plates. 

The features introduced in LISPER's design, including the C-shaped braces and the meshes, are mainly designed to constrain the undesired radial elongation of the silicone rubber's chamber and consequently increase the response rate. The bellow structure helps realize high linearity between input pressure and bending angles or output forces. It also provides a higher range of motion and output forces. The quantitative details will be in the analytical modeling section of LISPER.

\subsection{Mechanical Design: Scallop-shaped Pneumatic Robot (SCASPER)}
SCASPER (Fig. \ref{fig_2}(f)-(i)), inspired by scallop fan shells, was designed to generate a large output force/torque and high linearity bi-directional motion. SCASPER is classified as an inextensible airbag pneumatic actuator, which usually generates larger moments of force (2 Nm to 15 Nm) than silicone rubber-based soft robots like LISPER. However, conventional air-bag-based actuators predominantly suffer from (1) low linearity between inflation pressure and bending angles, (2) slow deflation speed, (3) severe hysteresis for bi-directional translation, and (4) complicated fabrication processes. These issues were considered in the development of SCASPER.

The design of SCASPER features a rotationally patterned arrangement of rectangular airbags, with two external corners trimmed off. This modification brings the contact points of the bags closer to the rotational axis, thereby enhancing the extension angle ${\theta}$ (Fig. \ref{fig_2}(f)). To speed up deflation and enhance the linearity between pressure and angle, silicone rubber strips were added to connect the external edges of the airbag (Fig. \ref{fig_2}(i)). These strips facilitate SCASPER’s tendency to return to its original position. A notable feature of SCASPER’s mechanical design is its construction from polyethylene heat-shrinking tubes, eliminating the need for layer stacking. The tube material does not impact performance because the angular extension does not rely on bag elasticity. Furthermore, this design allows for individual airbag airtightness checks before assembly into the overall structure. 

\subsection{Finite Element Analysis of LISPER and SCASPER}
In this section, we discuss the application of FEA to simulate the relationship between inflation air pressures and corresponding bending angles for LISPER and SCASPER, considering their complex geometric structures. The simulation serves three primary purposes: (1) to provide proof of concept regarding bending direction, expected deformation,  the linearity between air pressure and bending angles, and the output force/torque in different angular constraints; (2) to enable comparison between simulation results and experimental data; and (3) to identify stress concentration areas that could be potential sources of significant air leakage.

\begin{figure*}[ht]
    \centering
    \includegraphics[width=3.2in]{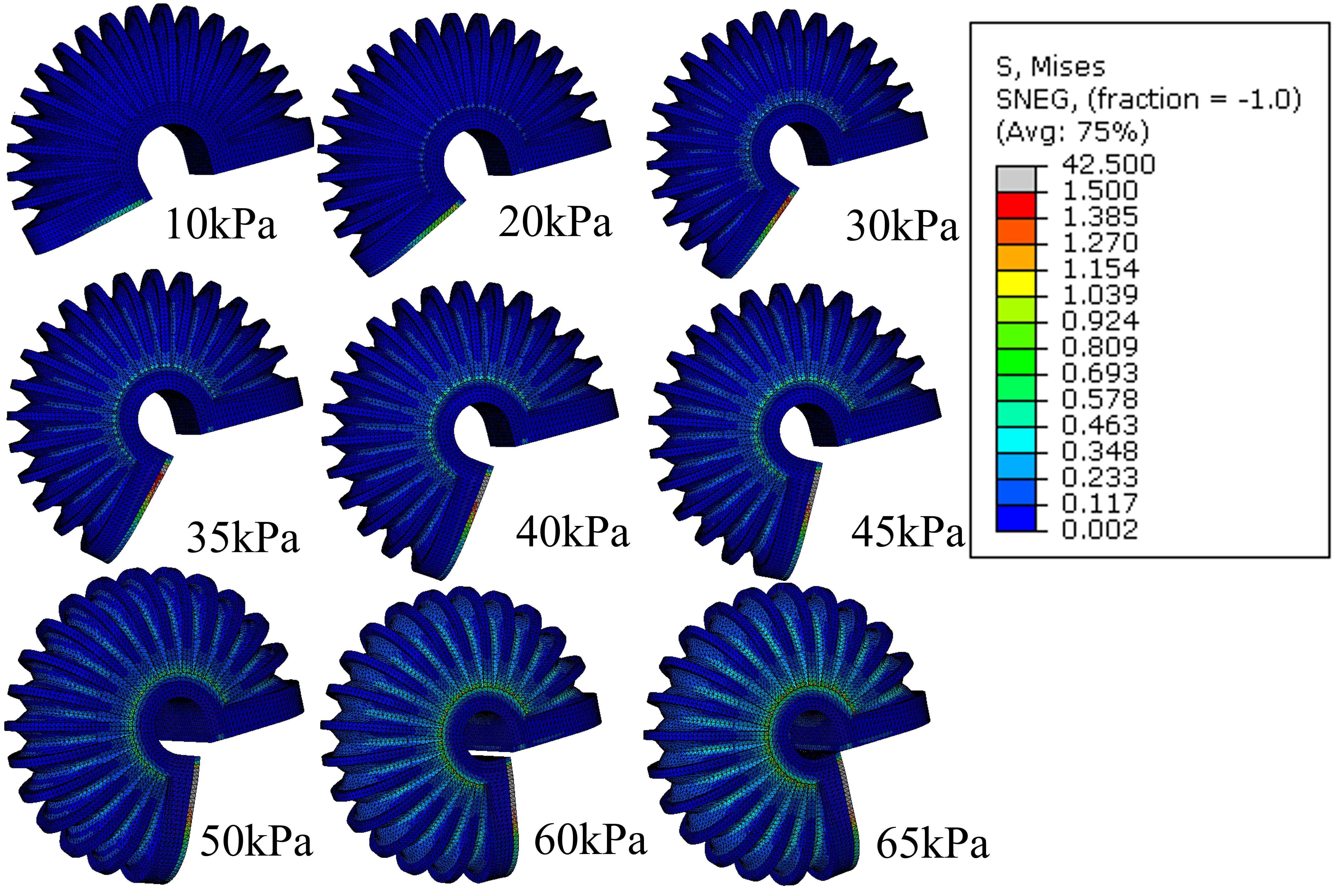}
    \caption{FEM simulation of the stress distribution of LISPER with pressure range from 10 kPa to 65 kPA. Note: The c-shaped brace is hidden on the image for clear demonstration.}
    \label{fig_4}
\end{figure*}

For the free motion of the simulation for LISPER, we assumed that the brace lockers and C-shape braces have large stiffness and do not deform during simulation by applying an elasticity of 4.4 MPa and Poisson's ratio of 0.22 based on the experiment conducted by Wang et al. (2019) \cite{wang2019effect}. The mesh introduced in Section II is excluded from the simulation, as the inverse folding does not occur in the unloading state. A curved 2D plane was added at the middle plane of the TPU to represent the neutral layer. The Yeoh model was chosen for its simplicity and minimal parameter requirements in describing the hyper-elastic model. A general static model was used, as the inflation can be considered a quasi-static process. The inflation pressure was applied uniformly inside the internal chamber, increasing linearly from 0 kPa to 100 kPa in increments of 5 kPa (Fig. \ref{fig_4}). The simulated relationship between pressure and angles, along with a comparison with the experimental results, is shown in Fig.  \ref{fig_8}(a).

\begin{figure}[htb]
\centering
\includegraphics[width=3.3in]{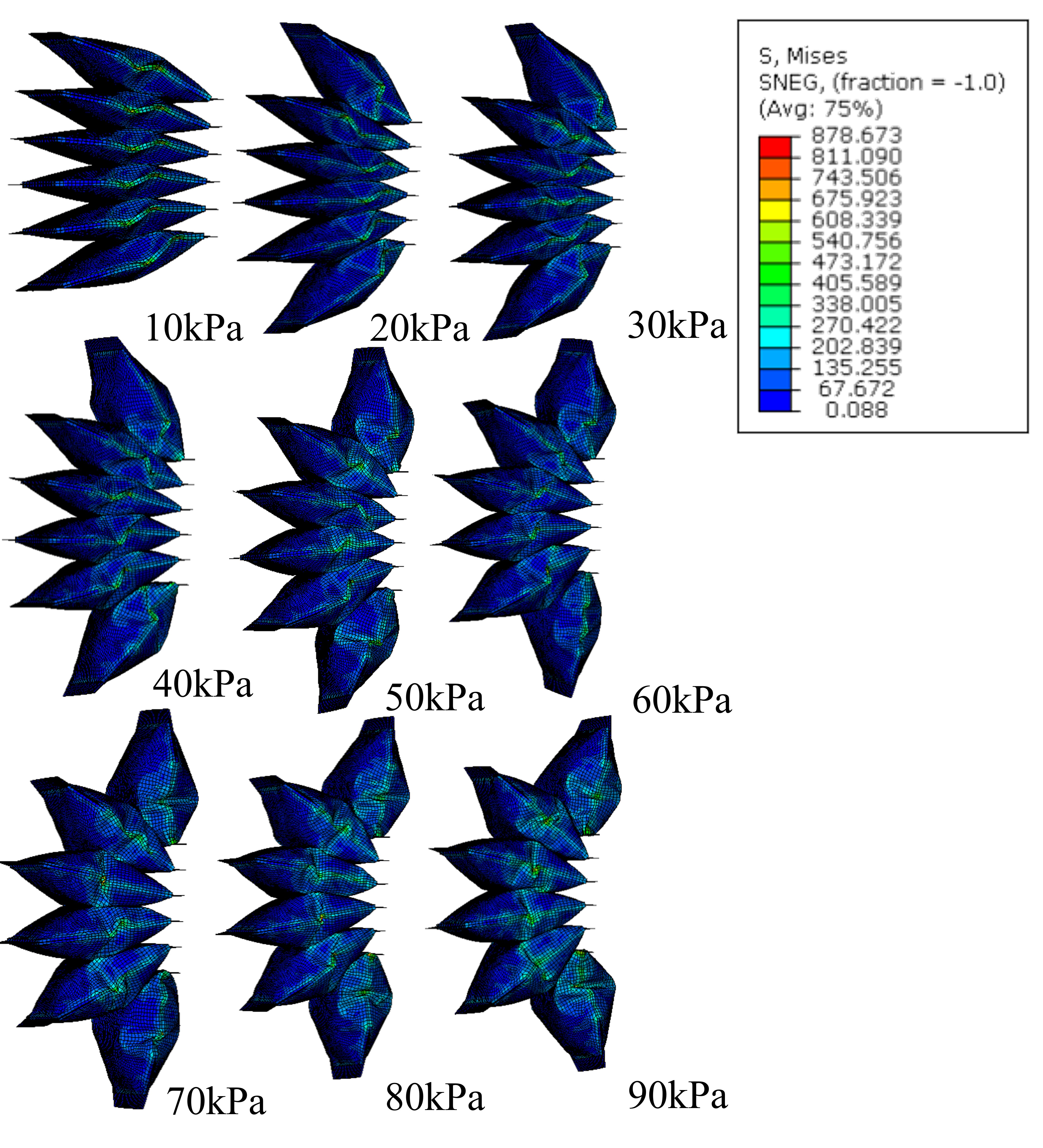}
\caption{FEM simulation of stress distribution of SCASPER with pressure range from 10 kPa to 90 kPA}
\label{fig_5}
\captionsetup{justification=centering,margin=0.5in,font=Large, labelfont=Large}
\end{figure}

The free motion of the simulation of SCASPER is based on the assumption that the material used for the airbag is inextensible. We set the young's modulus as 6.5 MPa and Poisson's ratio to 0.02. These values were not obtained through rigorous experimentation, as the simulation results are not significantly affected by the elasticity properties of the airbag itself  (Fig. \ref{fig_5}). Here, we applied dynamic explicit simulation. To determine the moment when all airbags attach, we applied the set pressure as a step input and allow the simulation to run for a sufficient amount of time. The simulated extension angles tend to oscillate in a sine wave pattern, from which we found the mean value (the offset of the sine wave) and defined it as the extension angle in a stable state. From Fig. \ref{fig_5}, it is evident that the high stress is primarily distributed around the two side surfaces of the airbags, which explains the choice of a shrinking tube instead of two separate fabrics for the airbag construction. The simulated relationship between pressure and angles is shown in Fig. \ref{fig_8}(b), along with a comparison with the experimental result. 

\begin{figure}[htbp]
      \centering
      \includegraphics[width=3.2in]{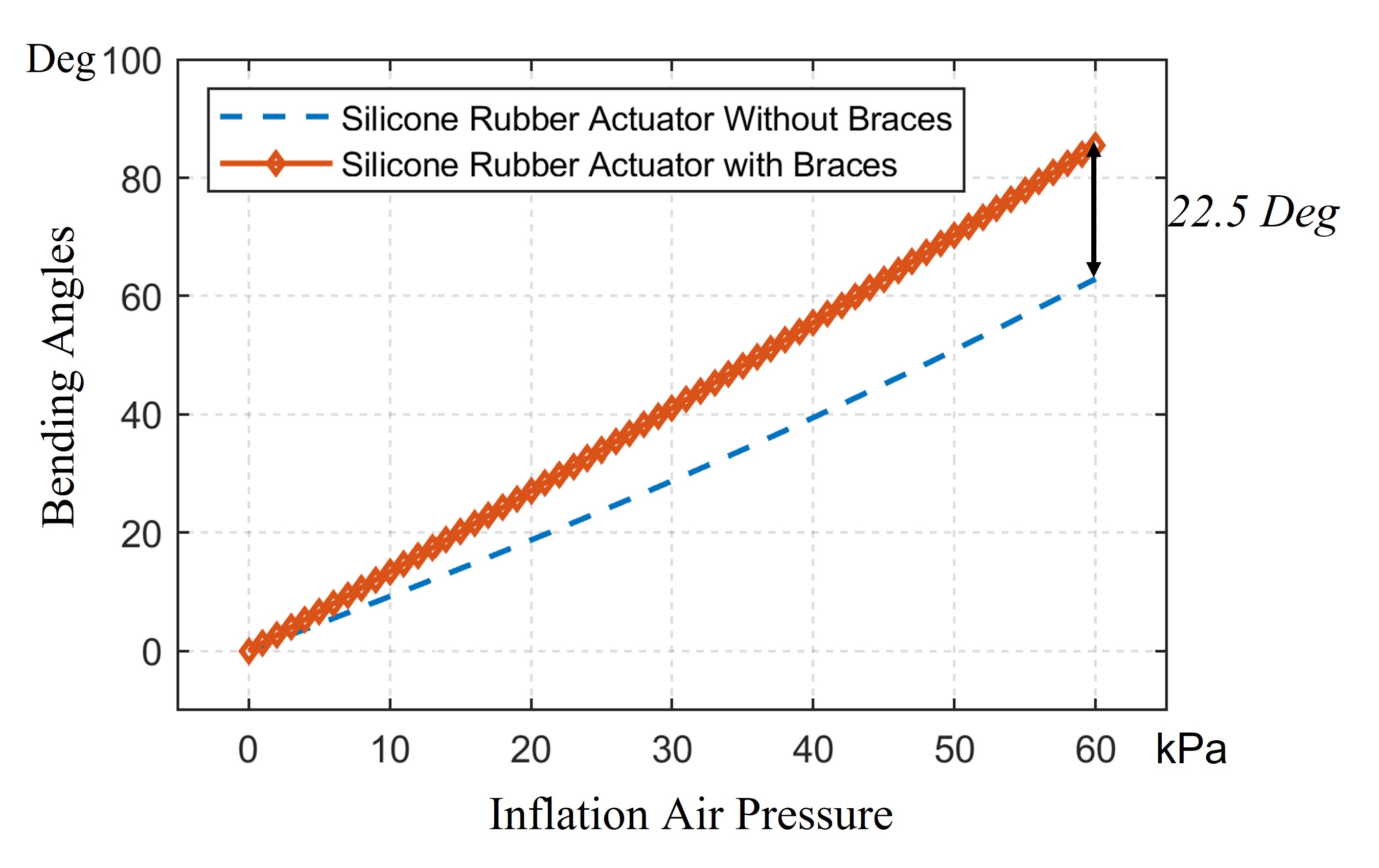}
      \caption{The comparison in the simulation of LISPER with and without c-shaped braces. }
      \label{fig_brace}
    \end{figure}

\begin{figure*}[htbp]
      \centering
      \includegraphics[width=3.2in]{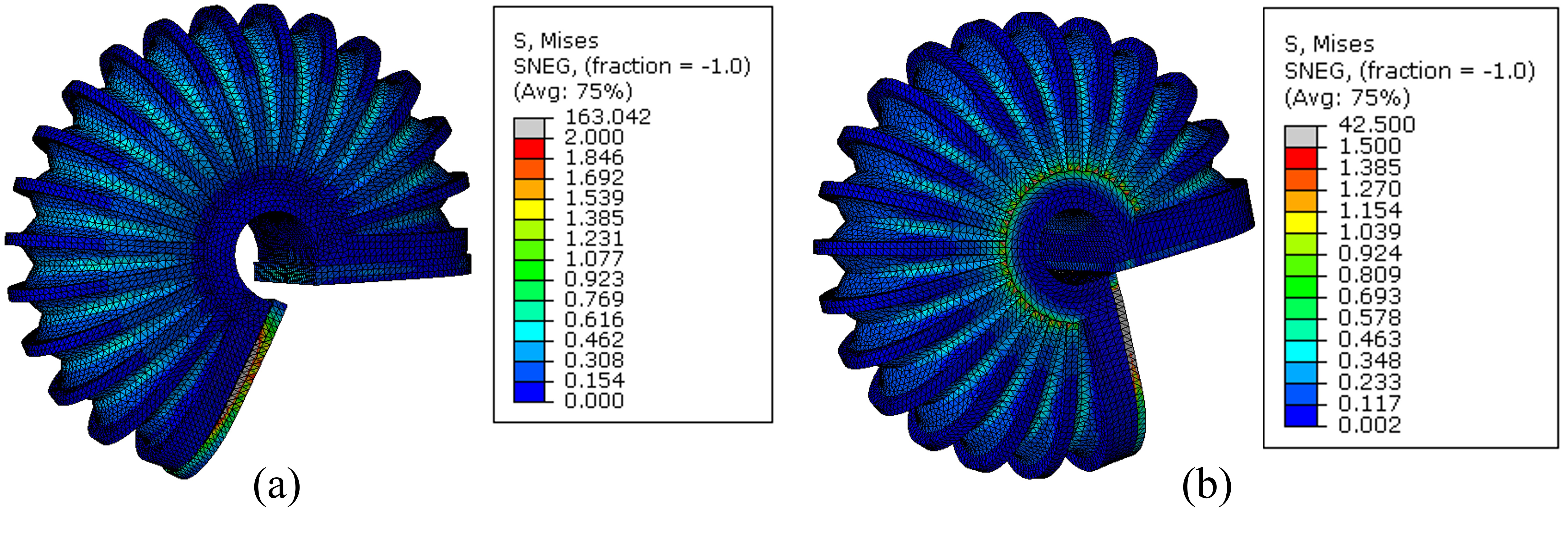}
      \caption{The comparison of LISPER actuator with or without braces at 65 kPa pressure input. The figure (a) is LISPER without braces and the figure (b) is with braces.}
      \label{fig_fea}
    \end{figure*}

We also verified the impact of the C-shaped braces and brace locker rings, which were added to the external edges of each fold. Through FEA analysis, we found that the C-shaped braces and brace locker rings increased the bending angle by 22.5° at 60kPa Fig. \ref{fig_brace} and Fig. \ref{fig_fea}.

To estimate the output force/torques in certain angles, we fixed multiple flat plains to constrain the motion of the LISPER and SCASPER in 0 deg, 45 deg, and 60 deg. and adding the input pressure from 0 kPa to 100kPa. The normal reaction forces from the plains are defined as the output forces/torques from the actuators. The simulation result is shown in Fig.\ref{fig_8} (b) and (e). 

\subsection{Fabrication and Assembly of LISPER} 
\textbf{LISPER} features a silicone rubber body made of Tin cure silicone rubber. After comparing Shore A hardness levels 10, 20, and 40, \textbf{20A} (Smooth-on Mold Max 20) was chosen for its ability to facilitate smooth bending with moderate pressure variation in both bending and extending directions. LISPER’s construction included C-shaped braces and corresponding locker rings, two 3D printed PLA bottom plates (Polymaker PLA Filament) sealing the internal chamber ends, and an air inlet attached to one of the plates. The inlet hole was connected to a PU tube (Beduan Pneumatic Air Tubing Pipe) with epoxy (J-B Weld Pro Size ClearWeld 5 Minute Set Epoxy). Each brace housed four squared holes for positioning fabric meshes (50” 8600 Polyester Dive Mesh, Seattle Fabric Inc.). The fabric mesh layers were placed between the braces and the silicone rubber body. Fixtures for the mesh utilized a series of C-shaped small arc pieces that grip the bellows’ edge and one side of the mesh hexagon. The silicone body features 15 bellows, which necessitated 15 pairs of C-shaped braces and locker rings.  

A two-mold casting technique was employed to fabricate LISPER's silicone rubber body (Fig. \ref{fig_2}(e)). A primary fabrication challenge was ensuring the high-viscosity liquid silicone rubber mixture uniformly fills the entire mold’s empty space. Failed castings are often a result of pores concentrated on the thin walls of the silicone rubber body. Therefore, syringe injection is unsuitable for filling the molds with silicone, as the high-viscosity liquid mixture may initially fill spaces with lower flow resistance, trapping air in small spaces and causing severe pores across the silicone rubber body. To address this issue, we utilized a vacuum pump and then decreased the temperature during the curing process. Additionally, we introduced air vent patterns, material inlet patterns, and a 3D-printed funnel to effectively prevent the accumulation of pores. The degassed silicone rubber liquid was poured into the funnel after being cooled with ice water. Positioned on the material inlet pattern atop the upper mold, the funnel allowed the liquid to flow slowly to the bottom of the empty space, pushing air upward to the air vent pattern locations.

In the molds used to fabricate LISPER, the two cores within the molds created a hollow chamber inside the silicone rubber body (Fig. \ref{fig_2}(e)). As detailed in the mechanical design section, one side of the silicone rubber body’s internal surface features a bellow-shaped structure, which was achieved by utilizing bellow-shaped cores. Each core contained two rectangular plugs corresponding to two grooves on the molds, ensuring accurate core-mold positioning and preventing downward core bias due to gravity.

A 3D-printed TPU layer (OVERTURE TPU Filament) was inserted around the mold’s bottom, which was wrapped with silicone rubber post-casting and embedded into the silicone rubber body. The holes on the TPU layer’s surface improved silicone rubber and TPU adhesion.

The chamber sealing process involved attaching two 3D-printed plates to the actuator’s ends. Additional freshly mixed silicone rubber was then added to the chamber ends. This was done by orienting the actuator upward and placing the SIL-Poxy coated plates (SIL-Poxy-Silicone Rubber Adhesive, Smooth-on Inc.) horizontally on both ends. The newly added silicone rubber mixture was then applied, and once it cured, it effectively sealed the chamber and firmly attached to the rest of the silicone rubber body.

\subsection{Fabrication and Assembly of SCASPER}
The inextensible membrane-based SCASPER is composed of six airbags stacked together, maintaining a controlled distance between each one. Each airbag was crafted from a polyethylene heat-shrinking tube (ELECFUN 2in Heat Shrink Tubing), chosen for its compliance, wear resistance, accessibility, and affordability. The heat-shrinking tube's sides were pre-sealed, reducing the likelihood of air leakage. The entire heat shrink tubing was divided into six equal-length pieces, with 5-hole and 6-hole patterns drilled on the distal (furthest from the rotational axis) and proximal (closest to the rotational axis) sides of the actuator to fix the bolts and nuts. Two corners on the distal side of each airbag were removed to bring the contact points between each airbag closer to the rotational axis of the SCASPER. The inlet was connected via a PU tube, and four long, flat ABS plates were used to seal the distal and proximal sides and secure the PU tube Fig. \ref{fig_2}(h). 

The six airbags were aligned and stacked on top of each other using pipe positioning rings. The six inlet pipes were merged into one to connect to the solenoid valve. Silicone rubber strips (Smooth-On Ecoflex 00-20 Super Soft Silicone) were connected at the distal end of each airbag (Fig. \ref{fig_2}(i)), which enables the adjustability of SCASPER.

The fabrication process of SCASPER takes approximately 45 minutes, excluding individual airbag airtightness testing. Overall, the fabrication procedure is significantly simplified, not requiring laser cutters, sewing machines, or heat-sealing machines. Given the strong adhesion of polyethylene heat-shrinking tubes to most types of tapes, achieving airtightness is relatively easy. SCASPER can withstand 150 kPa pressure input without experiencing irreversible deformation or significant air leakage. 

\section{Analytical Modeling of LISPER and SCASPER}
This section introduces the mathematical deduction process of the analytical models of LISPER and SCASPER. In these models, we use the general form of \(F = f(P,\theta)\), where \(F\) is the output force for the actuator, \(P\) is the inflected pressure, and \(\theta\) is the bending angle. The focus of the modeling section will be on explaining the inflation process of the bellow structure. The verification of the two analytical models is elaborated in the Mechatronic System for Experiments subsection in the Experiments and Results section, where we compare the models to experimental results.
\begin{figure*}[hbtp]
\centering
\includegraphics[width=\textwidth]{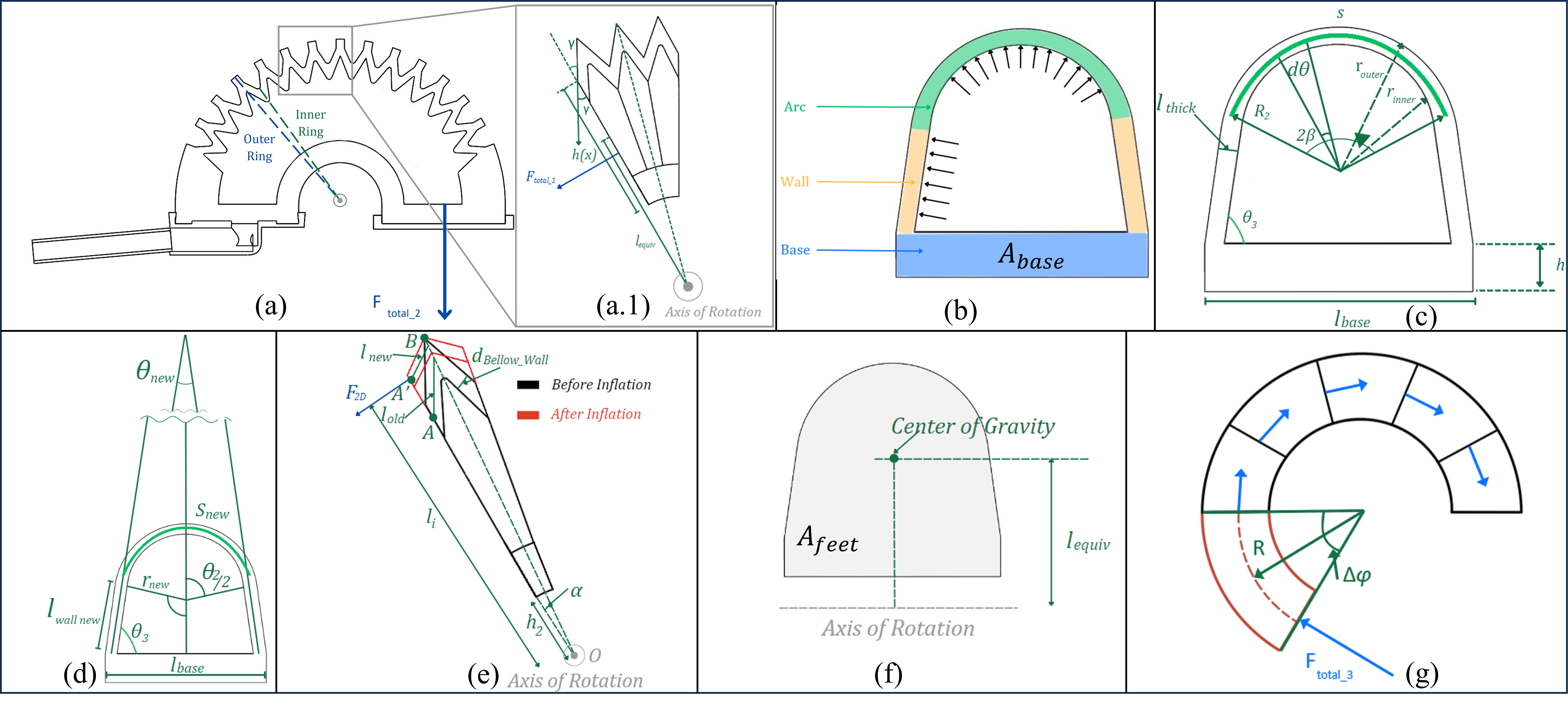}
\caption{Geometric diagram of LISPER. (a) The sectional view of silicone rubber body. The outer ring section is constrained by PLA rings, the inner ring is the smallest contour of each bellow segment. (a.1) The zoom-in view of three pieces of bellow segments. (b) The labeling of three sections of small ring, arc, wall, and base. (c) Dimension labeling of the inner ring before inflation. (d) Dimension labeling of the inner ring after inflation. (e) The folding deformation of the bellow structure before and after inflation. (f) Equivalent center of gravity and equivalent moment of arm. (g) The side view of the base section when it is bent.}
\label{fig_9}
\captionsetup{justification=centering, font=Large, labelfont=Large}
\end{figure*}
\begin{figure*}[hbtp]
\centering
\includegraphics[width=\textwidth]{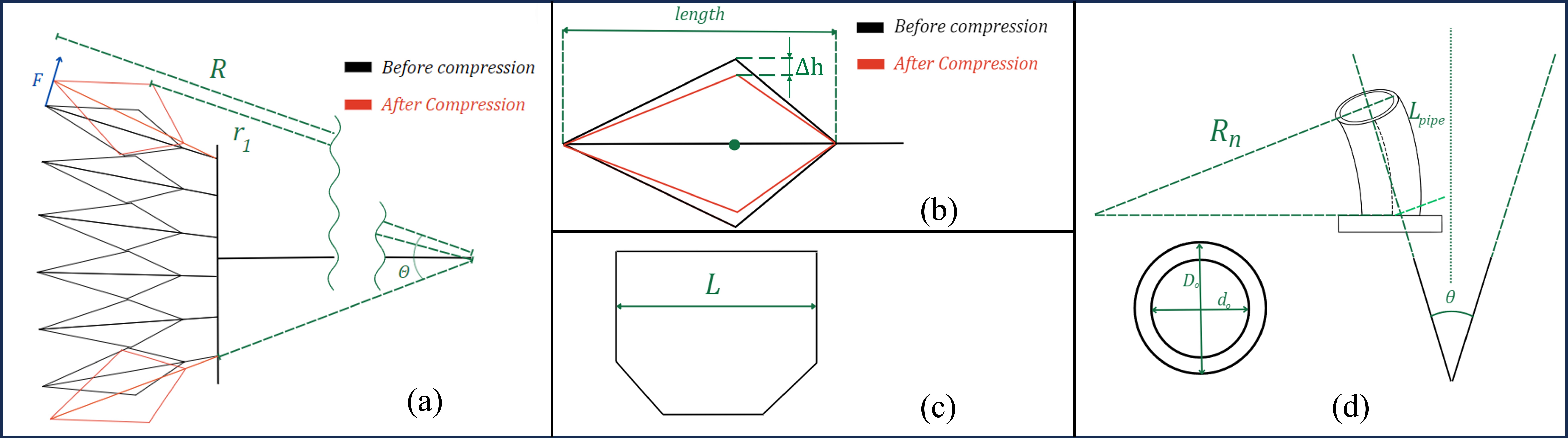}
\caption{Geometric diagram of SCASPER. (a) The geometric labeling of SCASPER before and after compressed. F is the force output r1 is the moment of arm from the contact point between airbags to the center of rotation. (b) The sectional view of one airbag before and after compression from the environment. (c) The width of each airbag from the top view. (d) The geometric labeling of the PU pipe when they are bent.}
\label{fig_10}
\captionsetup{justification=centering, font=Large, labelfont=Large}
\end{figure*}
\subsection{LISPER Analytical Model: The Modeling of the Unfolding Process of the Bellow Structure}
The analysis of the unfolding process of LISPER is decomposed into three sections of the geometry of the soft robot structure: (1) the bellow structure, (2) the two bottom areas of the silicone rubber body, and (3) the arc structure.

Fig. \ref{fig_9} describes the diagrams complementing the following modeling. Fig. \ref{fig_9} (a), (a.1) and (b) show the common parts that will be mentioned throughout.

The modeling process is formed based on several assumptions: (1) the linearization of hyper-elastic materials, which is based on the observation that the strain on the silicone rubber body is small; (2) the outer contour of the bellow segment does not extend given the PLA ring constraint of its outward motion; (3) the silicone rubber section between point A and point B in Fig. \ref{fig_9}(a.1) is straight, based on experimental observation.

From \cite{marechal2021toward} which provides raw data for commonly used silicone rubber materials, we observe that when the range of strain is from 0 to 1,  the stress-strain curve shows a high linearity, where Young's modulus could be approximated by 
\begin{equation}
    E= \sigma/\epsilon = 1.53 MPa
\end{equation}
Where $E$ is  Young's modulus of the silicone rubber body, $\sigma$ is uniaxial stress, and $\epsilon$ is the strain of silicone rubber.

Using the Young's modulus obtained, we investigate the mechanical behavior of LISPER during inflation. Fig. \ref{fig_9}(c) shows a cross-sectional cut of one of the bellows of LISPER, where

$\theta_3$ - angle between the base of the bellow and the wall

$R_2$ - radius from the center of the bellow to the external ring

$h$ - height from the lower area of the base to the upper area of the base

$2 \beta$ - angle of the ring of the bellow (constant value of initial position)

$d \theta$ - differential angle of the ring of the bellow 

$l_{base }$ - length of the base

$s$ - arc length

Once the air pressure is increased, the arc elongates following the Poisson effect. The resulting differential arc length, $d s_{n e w}$, of the bellow during inflation can be expressed as:
\begin{equation}
\label{eq:arc_length}
    d s_{n e w}=d s_{initial }+d s_{elongated }=r d \theta+\frac{r  P l_{thick} d\theta }{E v}
\end{equation}
Since $d s_{initial }=r d \theta$ and $\displaystyle d s_{elongated }=\frac{r  P \Delta rd\theta }{E v}$

$P$ - pressure during inflation

$r$ - radius of the middle layer between the inner ring and outer ring expressed as $\displaystyle r=\frac{{r_{inner} }+r_{outer }}{2}$

$l_{thick}$ - wall thickness of the whole small ring

$d s_{initial}$ - initial arc length,  $d s_{initial}=r d \theta$

$d s_{elongated }$ - increase in arc length

${v}$ - Poisson's ratio

In application, we usually assume the Poisson's ratio of silicone rubber to be around 0.5. In our practice, we found slight variations of Poisson's ratio do not cause a visible impact on the numerical results.

Integrating Equation \ref{eq:arc_length} from $-\beta$ to $\beta$, we can find the elongated arc length of the bellow during inflation, $s_{n e w}$
\begin{equation}
\label{eq:arc_new}
    s_{new }=\int_{-\beta}^\beta (r+\frac{r_{new} P l_{thick}}{E v} )d \theta=\left(r+\frac{r_{new }Pl_{thick}}{E v}\right) 2 \beta
\end{equation}
where $r_{new}$ is the new radius of the middle layer during inflation.
Assuming that $\beta_{new} \approx \beta$, we can have:
\begin{equation}
\label{eq:arc_new2}
    s_{new }=r_{new } \cdot 2 \beta
\end{equation}
By substituting Equation \ref{eq:arc_new2} into Equation \ref{eq:arc_new}, this leads us to finding the new arc length:
\begin{equation}
\label{eq:new_arc3}
    s_{new }=2 \beta r+\frac{s_{n e w} \cdot l_{thick}\cdot P}{E v}=\frac{2 \beta r}{1-\frac{l_{thick} \cdot P}{E v}}
\end{equation}

To characterize the elongation of the side wall, consider Fig. \ref{fig_9}(b), which shows the pressure profile along the wall.  The elongation of the side wall can be found through:
\begin{equation}
\label{eq:new_wall}
    l_{wall\_new }=l_{wall\_initial }+\frac{l_{wall\_initial } \cdot l_{thick}\cdot P}{E v}
\end{equation}

$l_{wall\_initial }$ - length of the wall before elongation

$l_{wall\_new }$ - length of the wall after elongation.

Since we know $s_{new }$ from Equation \ref{eq:new_arc3} , ${l_{wall\_new} }$ from Equation \ref{eq:new_wall}, and ${l\_base }$ is a constant, we can find $\theta_2, \theta_3$ and $r_{new }$ by simultaneously solving Equation \ref{eq:three_3} using a numerical solver and with the help of auxiliary lines along the plane of the bellow's structure, as shown in Fig. \ref{fig_9}(d):

\begin{equation}
\label{eq:three_3}
\left\{\begin{array}{l}
\displaystyle \frac{\theta_2}{2}  = \theta_{3}\\
\displaystyle \theta_2 = \frac{l_{new}}{r_{new}}\\
\displaystyle \frac{l_{base}}{2\cos{\theta_{3}}} = l_{wall\_new} + r_{new} \displaystyle \tan{\left(\frac{\theta_2}{2} \right)}
\end{array}\right.
\end{equation}
After acquiring all the geometric parameters, we can find the height of the sectional contour of the chamber, $h(x) $ in Equation \ref{eq:height}:
\begin{equation}
\label{eq:height}
h(x)= 
\left\{
\begin{array}{@{}l@{\extracolsep{\fill}}l}
\displaystyle \tan \theta_3\left(x-\frac{l_{\text{base}}}{2}\right) & \\
\displaystyle -\frac{l_{\text{base}}}{2}<x<-l_{\text{base}}+l_{\text{wall\_new}} \cos \theta_3 & \\
\displaystyle \sin \theta_3 \cdot l_{\text{wall\_new}}-r_{\text{new}} \cdot \cos \frac{\theta_2}{2}+\sqrt{r_{\text{new}}^2-x^2} & \\
\displaystyle |x|< l_{\text{base}}-s_{\text{new}} \cdot \cos \theta_3 & \\
\displaystyle \tan (-\theta_3)\left(x-\frac{l_{\text{base}}}{2}\right) & \\
\displaystyle l_{\text{base}}-l_{\text{wall\_new}} \cos \theta_3<x<\frac{l_{\text{base}}}{2}
\end{array}
\right.
\end{equation}
Now, we will characterize the forces generated by LISPER. To find the forces at the side spike of each bellow, we refer to (Fig. \ref{fig_9} (e)). First, we need to characterize the lateral compression of the bellow in Equation \ref{eq:compress}:
\begin{equation}
\begin{aligned}
\label{eq:compress}
& F_{2 D}=A \cdot \Delta l \cdot E \\
& A= d_{Bellow\_wall}+\frac{\Delta l}{v} \\
& \Delta l=l_{\text {new }}-l_{\text {old }}
\end{aligned}
\end{equation}
where $F_{2 D}$ is the force generated by the compression of each bellow segment in 2D, $d_{Bellow\_wall}$ is the width of the bellow wall, and $\Delta l$ is the compressed length of bellow segment.

Applying cosine law, we can find the Equation \ref{eq;cosine}:
\begin{equation}
\begin{aligned}
\label{eq;cosine}
& l_{\text {old }}{ }^2=O A^2+O B^2-2 \cos \alpha \cdot O A \cdot O B \\
& l_{\text {new }}{ }^2=O A'^2+O B^2-2 \cos \alpha \cdot O A' \cdot O B \\
& O A=h_2+h(x)
\end{aligned}
\end{equation}
where $\alpha$ is the angle between two extension side line intercepting at rotational axis $O$, $\alpha = \theta_{Bending\_angle}/(2\cdot N)$.

From the equations above, we can find $F_{2 D}(x)$, and by integrating from $-\frac{l_{\text {base}}}{2}$ to $\frac{l_{\text {base}}}{2}$, we can find the compression force of the entire bellow structure, $F_{3 D}$ in Equation \ref{eq:force} :

\begin{equation}
\label{eq:force}
F_{3 D}=\int_{-l_{\text {base}} / 2}^{l_{\text {base}} / 2} F_{2 D} d x
\end{equation}

To find the force generated at the side normal to the segment line of the bellow, $F_{\text {total\_1 }}$ (Fig. \ref{fig_9}(a.1)), we can use Equation \ref{f_total1}:

\begin{equation}
\begin{aligned}
F_{\text{total\_1}} &= \frac{T_{\text{total\_1}}}{l_{\text{equiv}}} \\
&= \frac{l_i \times F_{3D, i}}{l_{\text{equiv}}} \\
&= \frac{F_{3D} \cdot l \cdot \cos(90^{\circ}-\gamma) \cdot 2N}{l_{\text{equiv}}}
\end{aligned}
\label{f_total1}
\end{equation}

$\gamma$ - angle between extension line of the tip of the bellow and the bellow's segment

$N$ - number of bellows

$l_{\text {equiv}}$ - length of the equivalent force's executing point to the axis of rotation

$l_i$ - moment arm from the axis of rotation to the bellow

Now we need to find $F_{\text {total }2}$, as shown in Fig. \ref{fig_9}(a). $F_{\text {total }2}$ is the force generated by the air pressure at the feet of the LISPER (Fig. \ref{fig_9}(f)) in Equation \ref{eq:forceTotal2}.
\begin{equation}
\begin{aligned}
\label{eq:forceTotal2}
F_{\text{total\_2}} &= A_{\text{feet}} \cdot P
\end{aligned}
\end{equation}

$A_{\text{feet}}$ - Area of the feet of the actuator.

Now, we need to find the opposite force generated by the inner arc during the expansion of the actuator, $F_{\text {total\_3}}$ (Fig. \ref{fig_9}(g)) with Equation \ref{eq:forceTotal3}:
\begin{equation}
\begin{aligned}
\label{eq:forceTotal3}
F_{\text {total\_3}}=\Delta \phi A_{base} E R
\end{aligned}
\end{equation}
$\Delta\phi$ - difference of bending angle 

$\Delta\phi= \theta_{Bending\_angle} - \theta_{Initial\_Bending\_Angle}$

$A_{base}$  -  sectional area of the base

$R$ -  length of the equivalent forces from the geometric center of the sectional area to the rotational axis

The overall output force can then be expressed in Equation \ref{eq:dyn_lisper} as:
\begin{equation}
\label{eq:dyn_lisper}
F_{\text{output}} = \frac{\vec{l}_1 \times \vec{F}_{\text{total}\_1} + \vec{l}_2 \times \vec{F}_{\text{total}\_2} + \vec{l}_3 \times \vec{F}_{\text{total}\_3}}{l_{\text{equiv}}}
\end{equation}

Where $\vec{l}_1,\vec{l}_2,\vec{l}_3$ are moment of arm with respect to  $\vec{F}_{\text{total}\_1}, \vec{F}_{\text{total}\_2}, \vec{F}_{\text{total}\_3}$. We address $\vec{l}_1 = \vec{l}_2 = \vec{l_{\text{equiv}}} $ and $\vec{l}_3 = \vec R$.
Equation \ref{eq:dyn_lisper} describes the relationship between pressure input $P$, bending angle $\theta_{Bending\_angle}$ and output force $F_{output}$. Geometrically, the $F_{output}$ is proportional to the number of bellow segment $N$, and the thickness of bellow segment  $l_{thick}$. By setting $F_{output}$ as part of the cost function, geometrical optimization of the silicone rubber dimensions is realizable.

 Considering the contribution of $F_{\text {total\_1 }}$ quantitatively to the total exerted force by percentage in Equation .\ref{eq:percent},
\begin{equation}
\label{eq:percent}
    Perc_{Bellow\_force\_contribution}=\frac{F_{\text {total\_1 }}}{F_{output}} \times 100\%
\end{equation}
we found the contribution of force generated by bellow is around 35\%, which means the bellows structure plays an essential role in providing force out. The larger force output refers to the large range of motion because it provides a larger power supply to compensate for the inverse force (${F_{\text {total\_3 }}}$) and the resistance force from the environment.

\subsection{SCASPER Analytical Model: The Modeling of the Unfolding Process of the Bellow Structure}
To characterize the quasi-static behavior of SCASPER, we first need to find the relationship between the total angle of expansion and the inflation pressure. Then we will find the total force exerted by SCASPER. All necessary diagrams for SCASPER's modeling are outlined in Fig. \ref{fig_10}.

To find the relationship between the total angle of expansion of SCASPER and the pressure, we need to consider that the airbag is made out of non-extensible material and SCASPER inflates to an irregular shape. FEA simulation results were used for polynomial regression to find the angular extension as a function of pressure when there is no external loading and the airbag is not compressed as Equation .\ref{eq:polyfit}:
\begin{equation}
\label{eq:polyfit}
\Theta(P)=\operatorname{polyfit}(P)=0.0145 P^2+3.0507 P-1.1438
\end{equation}

$\Theta$ - total extension angle of the actuator

$P$- inflation pressure

Referring to Fig. \ref{fig_10}(a)-(b), when the airbag is compressed the work exerted from the environment will be converted to the change of  volume of the airbag. Therefore we can characterize the volume of one airbag in Equation \ref{eq:compress2} as
\begin{equation}
\label{eq:compress2}
\Delta V P=\Delta W
\end{equation}

$\Delta V$ - total volume compression on airbags

$\Delta W$ - total work exerted on the actuator from environment

Assumıng that $\Delta V \propto \Delta \Theta$, the work conversion can be expressed as Equation \ref{eq:conservative}

\begin{equation}
\begin{aligned}
& \Delta V P=F \cdot \Delta \Theta \cdot R =\Delta \Theta \cdot \tau
\end{aligned}
\label{eq:conservative}
\end{equation}

$K$ - equivalent spring coefficient  

$F$ - force exerted to the environment

$\tau$ - torque exerted by SCASPER 

$\Delta \Theta$ - angular difference before and after compression

We can then set
\begin{equation}
\begin{aligned}
\Delta V=\int_0^L \frac{1}{2} \text { length } \cdot 2 \Delta h d L
\end{aligned}
\label{eq_2}
\end{equation}
Plugging in Equation \ref{eq_2} in Equation.\ref{eq:conservative}, we get the relationship between the torque and pressure applied in Equation \ref{eq:tau_equation}:
\begin{gather}
\tau = \frac{\Delta V P}{\Delta \Theta} = \frac{\frac{1}{2} \cdot \text{length} \cdot L \cdot \Delta \Theta \cdot r_1 \cdot P}{\Delta \Theta} \notag \\
= \frac{\text{length} \cdot L \cdot r_1 \cdot P}{2}
\label{eq:tau_equation}
\end{gather}

Here we found an important conclusion that the exerted torque $\tau$ is irrelevant to $\Delta \Theta$. However, since the PU pipe would resist the extension of SCASPER when inflated, we assume the PU pipes are purely bent (Fig. \ref{fig_10}(d)) and apply the Euler–Bernoulli bending Equation \ref{eq:bending} to find the internal moment of the pipe:
\begin{equation}
\label{eq:bending}
M_{\text {pipe }}=\left(M_1+M_2+M_3\right)=2\left(\frac{E I}{R_1}+\frac{E I}{R_2}+\frac{E I}{R_3}\right)
\end{equation}

$M_n$ - internal moment in each PU pipe

$E$ - Young's modulus of PU pipe

$R_n$ - radius of the pure bending on each pipe

$I$ - torsion moment of inertia, $\displaystyle I=\frac{\pi\left(D_2^4-D_1^4\right)}{32}$, $D_1,D_2$ are the internal and external diameters of pipes

$R_n$  of each pipe can be described as:
$$
\begin{aligned}
& R_1=\frac{L\_pipe}{\Theta / 6} \\
& R_2=\frac{L\_pipe}{2 \cdot \Theta / 6} \\
& R_2=\frac{L\_pipe}{3 \cdot \Theta / 6}
\end{aligned}
$$

or more broadly:
$$
R_{N / 2}=\frac{L\_pipe}{N / 2 \cdot \Theta_{\text {real }} / N}
$$

where $N$ is the number of bags.

Therefore, the total torque produced by SCASPER is $M_{\text {total }}=\tau-M_{\text {pipe }}$.
This $M_{\text {total }}$ describes the relationship between pressure input, bending angle, and torque output.

\section{Experiments and Results}
\begin{figure*}[hbtp]
\includegraphics[width=\textwidth]{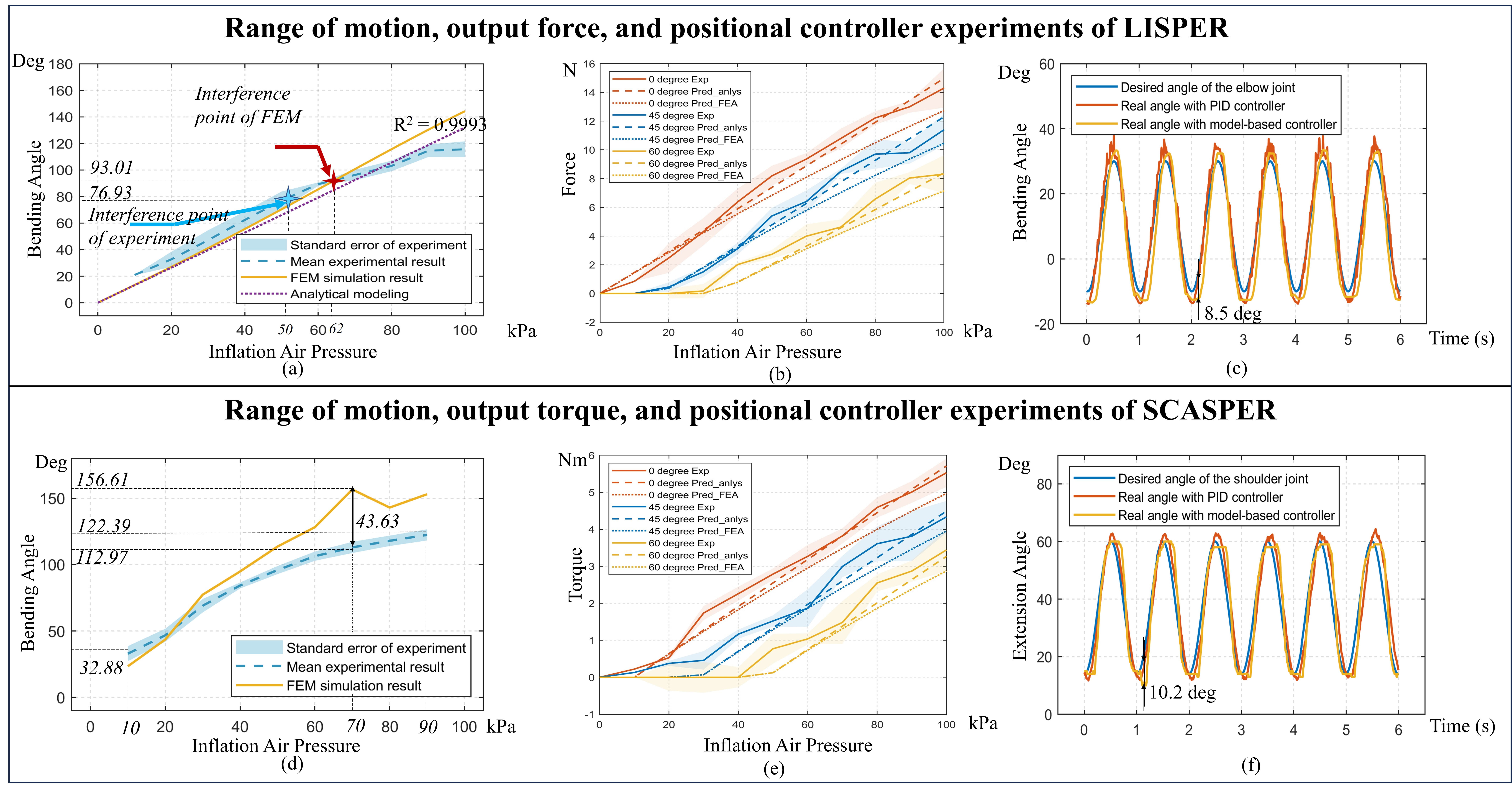}
\caption{(a) and (d) Comparison between modelings and experimental bending angle for LISPER and SCASPER, respectively. (b) and (e) Comparison among analytical model-based prediction, FEA, and experiment on Pressure vs Force and  Pressure vs Torque of LISPER and SCASPER under different fixed angles. (c) and (f) compare the PID model-free controller and the model-based position controller applied on the elbow and shoulder respectively.}
\captionsetup{justification=centering, font=Large, labelfont=Large}
\label{fig_8}
\end{figure*}

A comprehensive set of experiments was conducted to assess the performance of LISPER and SCASPER, focusing on aspects ncluding range of motion, force/torque versus pressure across various constrained angles, response rate, and fabrication complexity. The differences between the two actuators are detailed in Fig. \ref{fig_8}. Notably, SCASPER, owing to its inextensible layer properties, is capable of generating greater output force/torque, whereas LISPER offers reduced latency, high linearity, and relatively large output force/torque compared with previous studies. These properties play important roles in assistive motion around the shoulder and elbow.

\begin{table*}[ht]
\centering
\caption{Bandwidth-Related Properties of LISPER and SCAPSER}
\begin{tabularx}{\textwidth}{|X|X|X|X|}
\hline
\textbf{Condition} & \textbf{Range of Motion} & \textbf{Mean Time Error} & \textbf{Maximum Angular Error}  \\ \hline
1Hz Set Angle Input & 78.02° & 0.42 s & 17.14° \\ \hline
0.5Hz Set Angle Input & 81.46° & 0.25 s & 7.9° \\ \hline
0.25Hz Set Angle Input & 82.83° & 0.19 s & 6.85° \\ \hline
\end{tabularx}
(a) Range of Motion, Time Error, and Mean Error Angle of LISPER.
\vspace{5mm} % Adjust space here as needed

\begin{tabularx}{\textwidth}{|X|X|X|X|}
\hline
\textbf{Condition} & \textbf{Range of Motion} &\textbf{Mean Time Error} & \textbf{Maximum Angular Error} \\ \hline
Set Angle Input 1Hz & 89.53° & 0.42s & 32.54° \\ \hline
Set Angle Input 0.5Hz & 90.12° & 0.31 s & 29.74° \\ \hline
Set Angle Input 0.25Hz & 98.25° &  0.34 s & 21.54° \\ \hline
\end{tabularx}
(b) Range of Motion, Time Error, and Mean Error Angle of SCASPER without Rubber Pattern.

\vspace{5mm} % Adjust space here as needed

\begin{tabularx}{\textwidth}{|X|X|X|X|}
\hline
\textbf{Condition} & \textbf{Range of Motion} & \textbf{Mean Time Error} & \textbf{Maximum Angular Error} \\ \hline
Set Angle Input 1Hz & 87.33° & 0.40 s & 29.22° \\ \hline
Set Angle Input 0.5Hz & 82.32° & 0.38 s & 25.45° \\ \hline
Set Angle Input 0.25Hz & 81.55° & 0.30 s & 17.56° \\ \hline
\end{tabularx}
(c) Range of Motion, Time Error, and Mean Error Angle of SCASPER with Rubber Pattern.
\label{table:tab3}
\end{table*}

\subsection{Mechatronic System for Experiments}
The mechatronic system of the actuators was designed for measuring bending angles, controlling the air pressure loaded onto the actuators, and measuring the generated force. It also served to coordinate between two actuators to drive a two-degree-of-freedom human dummy arm. The system utilized a compressed pressure regulator (Hromee Compressed Filter Regulator Combo, Rohne Co Inc.) to stabilize the inflow air compression source at 200 kPa. The inflow pipe was further connected to a proportional electronic pressure regulator (ITV2050-212BL4, SMC Corp.) to control the input pressure into the actuator. The master microcontroller implemented here was the NI myRIO-1900 development board (National Instruments Corp.), which received orientation angles at 100 Hz from the NUCLEO-F091RC development board (STMicroelectronics Corp.) via UART ports. The orientation of the actuators was measured by two JY62 IMUs (Wit-motion Inc.) located at both ends of each actuator. The IMUs provided data about the linear acceleration, angular velocity, and angular orientation (in the form of Euler angles) to the NUCLEO board (Fig. \ref{fig_1}(a)-(b)). 

\subsection{Pressure vs Angle and the Comparison with Simulation}
To establish the relationship between angle and pressure, SCASPER and LISPER were mounted on an aluminum extrusion frame. Each was subjected to three inflation cycles, with pressures ranging from 10 kPa to 100 kPa for LISPER and from 10 kPa to 90 kPa for SCASPER, in increments of 10 kPa. The experimental results were then compared with FEM simulation results. As illustrated in Fig. \ref{fig_8}(a), the intervention point of the experiment and simulation is slightly misaligned, which could be caused by the outward offset of the bottom plates. However, the experimental data closely match the simulation results up to 50 kPa. Beyond the intervention point of FEM, the bending angle continues to increase as the intervention surfaces slide against each other, and the discrepancy grows. The maximum bending angle of LISPER is 112.2°. In Fig. \ref{fig_8}(d), the discrepancy between experimental and simulated extension angles increases after the pressure reaches around 30 kPa, reaching a maximum of 43.63° at 70 kPa. The overall discrepancy for SCASPER is large, which could be attributed to the stress from the deformed air tube attached to the end of each airbag. This design deficit will be fixed in the next version of the design. 

\subsection{Pressure vs Force/ Torque Relationship with Different Constraint Angles}
Force vs. pressure experiments under different bending angles were conducted for LISPER and SCASPER (Fig. \ref{fig_8}(b)-(e)), where the force was measured using a High Accuracy Digital Force Gauge (Omega Engineering Inc.). The actuators were constrained to 0 degrees, 45 degrees, and 60 degrees, and each case's measurements were conducted three times. Since LISPER's output was modeled using force, whereas SCASPER was modeled using torque, the experiment measured the output force of LISPER and the torque of SCASPER. The corresponding moment of force was then calculated by multiplying the force with the moment arm. The force was assumed to be exerted at the tips of SCASPER. The moment of the arm was measured from the tip of either LISPER or SCASPER to their rotational axis. The inflation pressure range goes from 10 kPa to 100 kPa for SCASPER and LISPER. The maximum force and moment for SCASPER is around 5.45 Nm, and for LISPER, it is around 11.5 N. 

\subsection{Bandwidth Analysis with Sinusoidal Set Angle Input}
\begin{figure}[hbtp]
\includegraphics[width=0.5\textwidth]{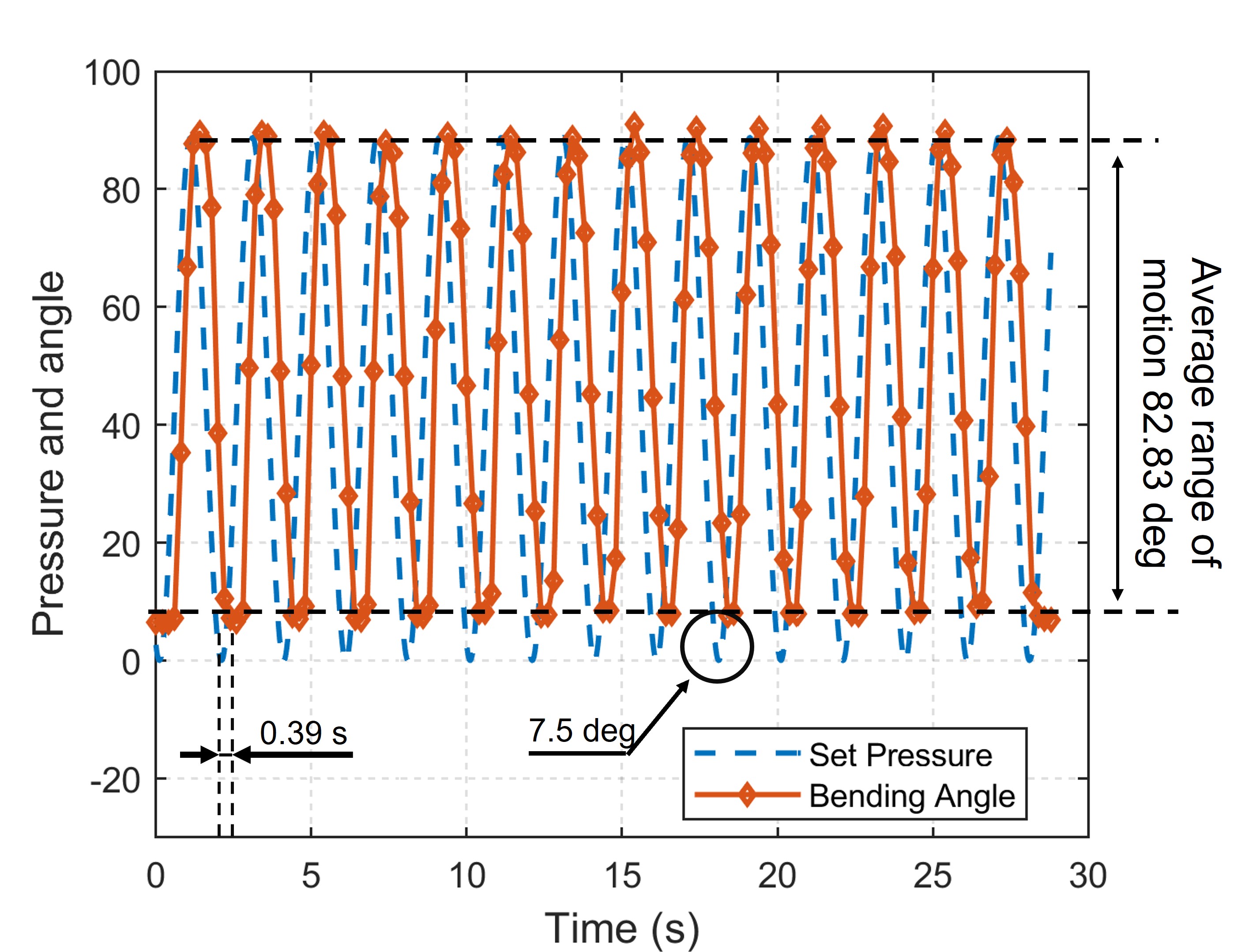}
\caption{ The experiment of set angle vs real angle under 4s sine set angle input of LISPER. }
\captionsetup{justification=centering, font=Large, labelfont=Large}
\label{fig_bandwidth}
\end{figure}

To find out how the rate of pressure change affects the dynamic range of motion of each unloaded pneumatic actuator, we added sinewave set pressure input to test the bandwidth of the actuators. The set angles desired are sine waves input range from 0° to 85°; the analytical models map from input pressure to desired angles are generated by experimental data from Fig. \ref{fig_8}(a) and (d) by a simple polynomial fit. The specific data are provided in Table. \ref{table:tab3}. For LISPER and SCASPER, the \textbf{Mean Time Errors} are around 0.3 s, which causes a negligible impact on patient-involved experiments. Given the elastic property of the LISPER body, the \textbf{Maximum Angular Error} of LISPER is much smaller than that of SCASPER. The elastic rubber pattern of SCASPER also shows its impact by decreasing the amount of \textbf{Maximum Angular Error} of SCASPER. One limitation we found from this experiment is that the angular errors for both actuators are generally large. This could be attributed to the limitation of the response rate of solenoid valves and the inefficient design of internal chambers. This deficiency will be solved in our further research with the help of more optimal designs for the internal chamber. One example of set angle vs. real angle for LISPER under 4-second periodic set angle input is demonstrated in the time domain in Fig. \ref{fig_bandwidth}, which shows that LISPER could follow the desired trajectory most of the time, except in the areas around the trough of the waveform. A similar situation happens to SCASPER. The severe follow-up error around low set angle input could be given to small elastic forces from the actuator bodies around the tough. SCASPER could further improve its follow-up capacity by increasing the elasticity of the strip rubber pattern. 

\subsection{Preliminary Feasibility Tests on Dummy Arm}
\subsubsection{Maximum Bending Angle Test}
To test the maximum bending angle driven by LISPER and SCASPER. The two actuators were positioned around the shoulder and elbow joints of a 2-DOF 3D-printed human arm to test the range of motion as shown in Fig. \ref{fig_1}(c),(d). To simplify the process, open-loop inflation pressures of 100 kPa and 90 kPa were set for LISPER and SCASPER, respectively. Based on the forward kinematics derived from joint angles (measured by the IMUs) to the end effect point, LISPER rotates 47.4 degrees under 80 kPa, and SCASPER rotates 54.4 degrees under 100 kPa. The end effect point had a maximum offset of 10.4 cm horizontally and 32.7 cm vertically. 

To further verify the feasibility of the two actuators, we set two control modes commonly used in upper-limb rehabilitation training position control and gravity compensation control. The position controller drives the two actuators to push the dummy arm to desired angles whereas the active motion of participants drives the gravity compensation mode of the two actuators. The elaboration on the two controllers will be in the Supplementary Material.

\subsubsection{Positional Controller and Gravity Compensation Controller}
In our experiments, the elbow and shoulder joints of a 2-DOF dummy arm were controlled by sinewave inputs to achieve desired angles ranging from -10° to 30° for the elbow and 16° to 60° for the shoulder. These input range sets reflect the physically constrained range of motion of the 2-DOF arm. The shoulder was driven by SCASPER, and the elbow was driven by LISPER. As depicted in Fig. \ref{fig_8} (c) and (f), the maximum deviation observed is approximately 8.5° at the elbow and 10.2° at the shoulder. Overall, the actual angles of both joints closely followed the set trajectories. The accompanying video demonstrates this synchronized sinewave movement. Notably, we observed relative errors around the turning points of the desired angles, likely due to the solenoid valves' limited response capabilities when handling oscillating reference pressures. 

Additionally, the gravity compensation controller was implemented based on the inverse quasi-static model and arm posture estimation. The aim was for the dummy arm to maintain three arbitrary positions, counteracting the gravitational pull. The gravity compensation test was conducted on the two joints independently and followed by a synchronized test on both joints.

\section{Discussion}
\begin{table*}[ht]
\centering
\caption{Comparison between designed actuators with other state-of-the-art works}

\centering
\begin{tabularx}{\textwidth}{|X|X|X|X|X|}
\hline
\textbf{State of The Art and LISPER} & Huichan Zhao Et al. & Kouki Shiota Et al. & Tze Hui Koh Et al.& LISPER \\ \hline
\textbf{Range of Motion (Deg) }& 85° (167kPa) & 90° & 90° & 112.2° \\ \hline
\textbf{Output Force(N)/ Torque (Nm) in 100 kPa} & 4.5N (150kPa) & 15 cNm & 12.2N & 12.5N/ 0.6 Nm \\ \hline
\textbf{Materials} & Silicone Rubber & Silicone Rubber, Fabric Strings & Silicone Rubber, Fabric Sheet, Polyethylene Heat-Shrinking Tubes & Rubber Stud Fastening\\ \hline
\textbf{Layers of Chamber} & Single-layer & Single-layer & Multi-layer & Single-layer \\ \hline
\end{tabularx}
(a) Comparison between LISPER and the state of the art.
\vspace{3mm} % Add space here

\centering
\begin{tabularx}{\textwidth}{|X|X|X|X|X|}
\hline

\textbf{State of The Art and SCASPER} & T. Abrar Et al. & Mae Irshaidat Et al. & Matthew A Robertson Et al & SCASPER \\ \hline
\textbf{Range of Motion (Deg)} & 135° & 112° & 42.5° & 122.5° \\ \hline
\textbf{Output Force(N)/ Torque (Nm) in 100 kPa }& 34.4N & 12N & 9N & 49.5N/ 5.5Nm \\ \hline
\textbf{Fabrication Device/ Materials} & Fabric Sheet, Rubber/ Sewing & PET, Butyl Rubber/ Sewing, Sealing & Fabric Sheet/ Heat Sealing Sewing & Polyethylene Heat-Shrinking Tubes/ Stud Fastening, Cellulose Taping \\ \hline
\textbf{Layers of Chamber }& Multi-layer & Multi-layer & Multi-layer & Single-layer \\ \hline
\end{tabularx}
(b) Comparison between SCASPER and the state of the art.

\label{table:tab2}
\end{table*}

\subsection{Comparison with Previous Work in Mechanical Design}
The mechanical design for the actuators focuses on improving the range of motion, force/torque versus pressure across various constrained angles, response rate, and fabrication complexity. For context and comparison, we reference previous works in this domain \cite{abrar2019epam,irshaidat2019novel,robertson2021soft,zhao2015scalable,shiota2019enhanced,koh2017design}, selecting studies that utilize similar materials (inextensible fabric or extensible silicone rubber), function as wearable robots and operate on pneumatic power. Detailed experimental outcomes are presented in Table \ref{table:tab2}. To standardize the comparison, we chose the range of motion and output force/torque at 100kPa. The output force/torque of Robertson et al. \cite{robertson2021soft} is calculated based on the linear range of motion and active linear stiffness. In Zhao et al. \cite{zhao2015scalable}, we chose 85° under 167Kpa as the direct pressure vs angle curve is not provided. A reading of 4.5N at 150Kpa was chosen for its highly nonlinear force/torque versus pressure curve. Overall, our actuators show equivalent or better performances than the referenced works. Additionally, our devices exhibit higher linearity than those in \cite{zhao2015scalable}. It is also noteworthy that the fabrication process, particularly the manufacturing equipment required for SCASPER, is significantly simpler compared to other fabric- or PET-based soft actuators.

In general, LISPER and SCASPER outperform all the previous works or show similar performance. We noticed the Maximum Angular Error in the response rate test across all three cases (refer to Table \ref{table:tab2}). The results demonstrate that the latency of pneumatic soft actuators, although partially mitigated by constraints on undesired extension and tunable elastic rubber patterns, still requires more advanced pneumatic systems and controller designs. With further geometric optimization, the range of motion and output force/torque can be further improved, the volumes of the actuators can be reduced, and the latency in the time domain can be attenuated. 

\subsection{Limitations of Analytical Model and Controllers}
The analytical model of LISPER assumes the linear property of \textbf{Smooth-on Mold Max 20} which is based on the observation that the strain exerted on the silicone rubber body is less than 1. Another simplification is we assume the side walls of the bellow structure will remain flat after inflation, whereas in reality, it is partially curved. 

In the analytical modeling of SCASPER, we neglect the irregular deformation of the inflated airbag and assume its cross-sectional view is triangular. The geometric parameters are set as constant and measured empirically. Apart from these, we also assume the friction between each bag is negligible. Although these assumptions can bring additional inaccuracy to the modeling, the position controller applied in this model follows up the desired trajectory accurately (Fig.\ref{fig_8}. (c) and (f)), considering we included a back-loop compensation. 

LISPER's accuracy is also proven in the experimental results of the force predictions based on analytical modelings \ref{fig_8}. (d). We noticed that in the force measurement experiment LISPER, the discrepancy between predicted force and desired force increases with pressure, which is caused by the sliding between the force gauge and LISPER.

Another assumption is the linearity between the stress and strain of the  \textbf{Smooth-on Mold Max 20}. Although the silicone rubber is commonly a hyperelastic material, from Fig. \ref{fig_8}. (a) it is easy to tell that the FEA simulation in free motion is a straight line, which provides support for the assumption of linear property in small-scale deformation. Another support of the linearity is from \cite{marechal2021toward}, where most silicone rubbers show linear stress-strain behavior within the strain range of 0 to 1. The team believes  this assumption on linearity is not the source of modeling inaccuracy.

The gravity compensation controller, essentially an open-loop system, enables the dummy arm to maintain its position under externally applied forces, thereby validating the precision of the analytical models. This controller's effectiveness is contingent on the accurate determination of the dummy arm's state values. However, it's important to note that this controller operates without back-loop error compensation, which means it does not actively correct for any deviations or errors that occur post-initial calibration. This aspect highlights a reliance on the initial accuracy of the system's state values and may suggest a potential area for enhancement in future iterations of the controller design.

\subsection{Fluctuation of Preliminary Test on Dummy Arm}
The evaluation of the controller's effectiveness was conducted using a 2-DOF (Degrees of Freedom) human dummy arm, as a substitute for human testing. This approach was chosen due to the challenges in discriminating the work contribution by humans from the actuators in a soft robot system.

A notable issue identified in the model controllers, as illustrated in Figure \ref{fig_8} (c) and (f) and the position controller in the attached video, is the fluctuation observed around the peaks and troughs of the sine wave, particularly for the elbow joint. This phenomenon may be attributed to the limitations inherent in the solenoid valves and the quasi-static models, which overlook the effects of velocity and acceleration. When compared to the model-free PID controller, it was observed that the elbow controller (Figure \ref{fig_8} (c)) exhibited increased oscillation. The shoulder PID controller demonstrated similar stability and accuracy as shown in Figure \ref{fig_8} (f), yet it also exhibited noticeable oscillation. These fluctuations could be caused by the overshot of the PID controllers rather than the design deficit of the mechanical design.

\subsection{Impact of Fabrication Design}
Our fabrication is different from many fabrications of the pouch-based structure. The fabrication design of SCASPER is largely simplified and requires only scissor cutting, 3D printed rigid frame, taping, and screwing. In contrast, many other pouch-based structures require sewing machines, heat-sealing machines, laser-cutting machines, etc. The fabrication time is largely decreased, given the concise structure design of the airbag stacks. The fabrication process of LISPER decreases the chance of air bubble trapping in complex geometric structures of the casting modes which is used to be challenging in large-size silicone rubber body fabrication. These modifications from traditional fabrication techniques guarantee the quality of soft actuators and improve the success rate of fabrication.

\subsection{Future Work: Human-Involved Experiments}
\begin{figure}
    \centering
    \includegraphics[width=0.5\linewidth]{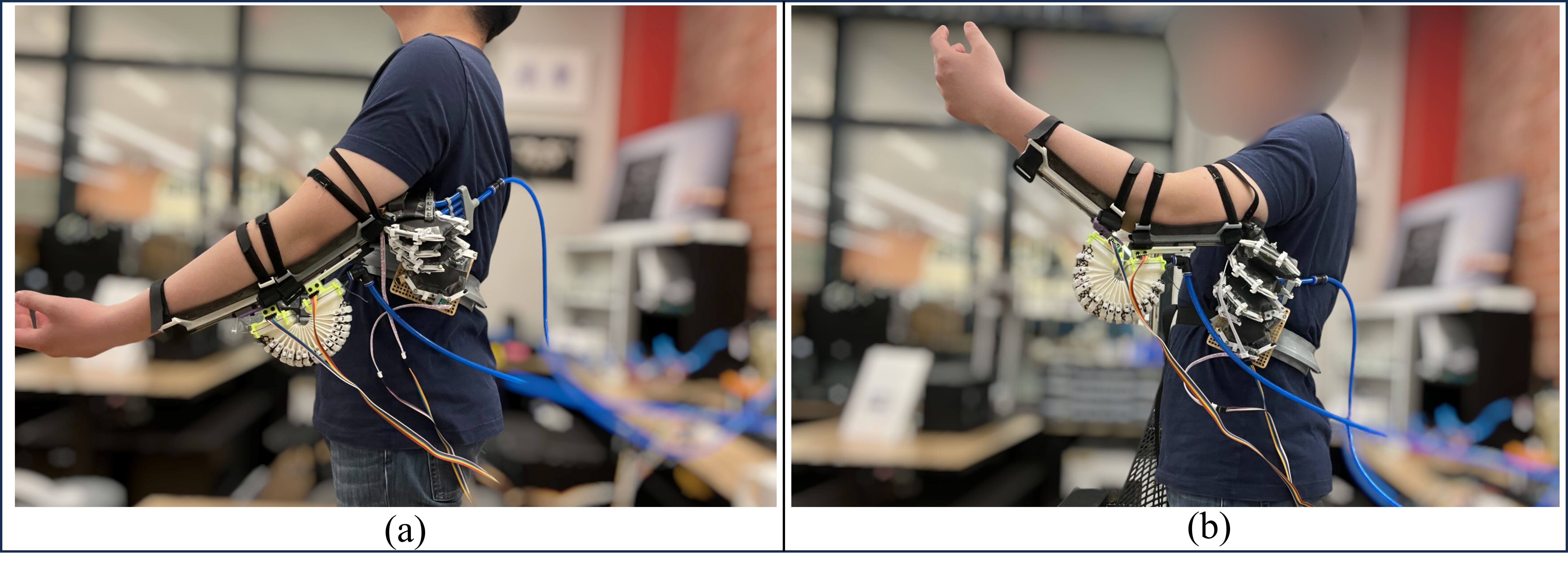}
    \caption{(a)-(b) shows the human subject practicing full arm extension and flexion wearing the two actuators.}
    \label{fig_11}
\end{figure}

In the future, the two actuators will be integrated into a complete single-arm exoskeleton. Several human-involved experiments will be conducted to verify the wearability and usability of the device. The group has developed the first version of an interface to hold the actuators on the human body, as shown in Fig. \ref{fig_11}. Preliminary tests on human subjects have indicated undesired sliding between the human body and the actuators. An optimal design should be implemented to address this undesired motion.  The next mechanical iteration shall focus on interface design to (1) have good alignments with human motion. (2) fit variations among different people considering their biomechanical properties. 

Other potential research directions, following the establishment of reliable usability and wearability, include: Conducting pilot tests for rehabilitation in a clinical setting. This would involve a paradigm composed of single-joint strength training and two-joint coordination training.
Developing dynamic modeling and control strategies that take into account the hysteresis of the actuators and the reduced dynamic model of the human upper limb.

\section{Conclusion}
In this study, we propose a new paradigm for designing bio-inspired pneumatic actuators for upper-limb rehabilitation, including the development of analytical models and fabrication processes. 
Furthermore, we conducted experiments to verify the dynamic properties of the two actuators and their coordinated performance on a mannequin arm. 
Specifically, we proposed the LISPER and SCASPER actuators for the elbow and shoulder, respectively. 
The LISPER actuator features detailed bellow-shaped folds, meshes, and braces, offering excellent performance in terms of range of motion, output force/torque, and linearity. In contrast, the SCASPER actuator is designed for time-efficient fabrication, with adjustable linear output force/torque achieved by modifying the silicone rubber strip pattern. It provides sufficient output force/torque to enable a wide range of motion and effectively support shoulder flexion.

The actuators were combined as a mechatronic system and deployed with position control and gravity compensation control mode to evaluate their practical performance. Although the system exhibited slight fluctuations at certain points of the working range, the two-degree-of-freedom system tracked the desired trajectories stably under position control and maintained the desired position under gravity compensation mode.

However, both soft actuators showed relatively large peak error values during high-speed repetitions, which can be attributed to limitations in pneumatic drivers, control strategies, and the geometric design parameters of the chambers. These limitations, along with other potential improvements discussed in the manuscript, will be addressed in our future research.

Our results indicate that this actuator design paradigm has significant potential for further development in future work. By refining and optimizing the design principles outlined in this study, researchers could enhance the functionality and versatility of pneumatic actuators for various applications, particularly in rehabilitation and assistive technologies. Future research could explore adapting these actuators for different joints or movements, integrating more advanced control strategies, and improving material properties to enhance durability and performance in real-world settings.

\bibliography{ref}

\end{document}

\onecolumn
\firstpage{1}

\title[Supplementary Material]{{\helveticaitalic{Supplementary Material}}}

\maketitle

\subsection{Analytical Model-based Position Controller Design}
\begin{figure*}[hbtp]
\centering
\includegraphics[width=0.95\textwidth]{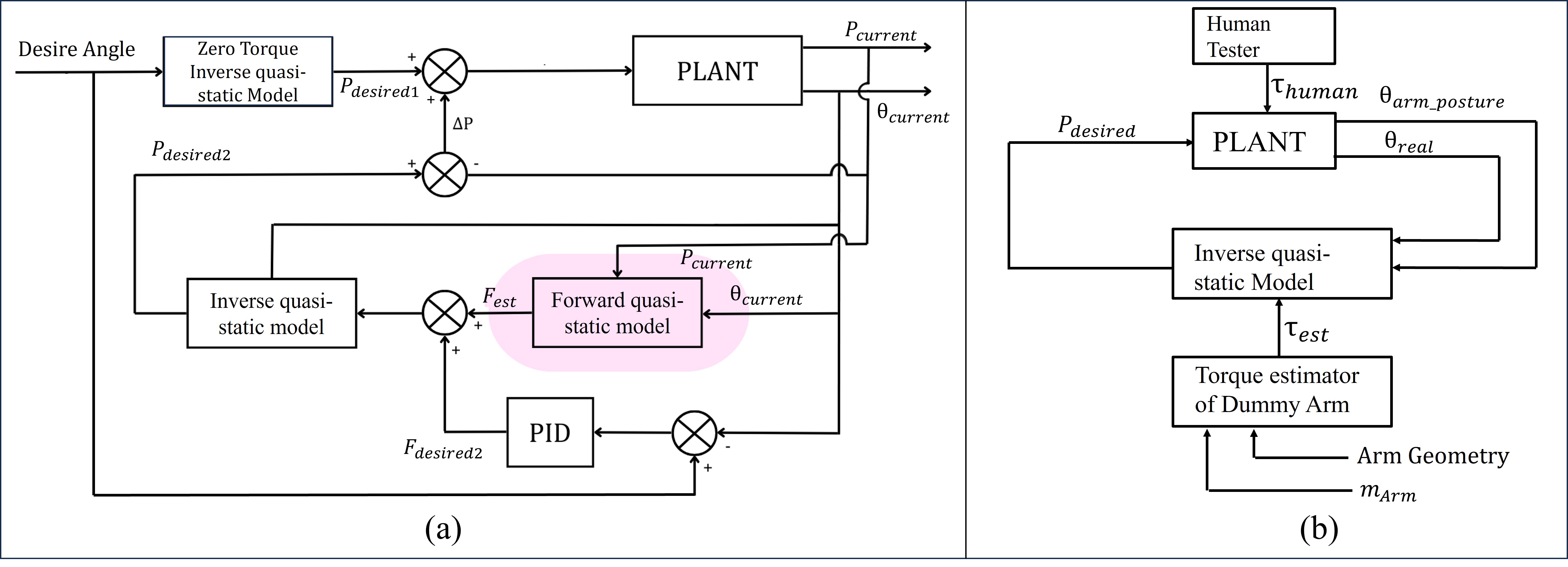}
\caption{(a) Schematic diagram of quasi-static model-based position controller. (b) Schematic diagram of gravity compensation controller}
\label{fig_12}

\end{figure*}

Utilizing the quasi-static models of LISPER and SCASPER, we developed a model-based controller for both systems (Fig. \ref{fig_12}(a)). By applying the model-based controllers, we enabled adaptive adjustment to varying loads on the actuators, which is a crucial aspect in clinical settings. This adaptability is essential due to the variability in limb weight and the effort exerted by patients, which can differ based on individual health conditions and inherent physiological differences. The framework was formed by the feed-forward zero torque inverse quasi-static model, where we assumed the external load to the actuators is zero. This model converts from the desired angle to the~$P\_desired1$ input. Considering there are unknown loads from the environment, the real angle would differ from the desired angle. Therefore, we applied a PID controller to add the desired force to compensate for the error angle. However, since forces were already applied to the environment by the actuator, we introduced a forward quasi-static model as the force estimator to estimate the force applied by the actuator in real-time. The summation of $F\_estimated$ and $F\_desired$ are placed in the inverse quasi-static model, generating another desired pressure $P\_desired2$. The $\Delta P = P\_desired2 - P\_current$ were then summed with $P\_desired1$ and given to the actuators.

In the experimental section, the desired angle was given as a sinewave in the system. Furthermore, in the video attached, we provide the full range motion of both the actuators and their synchronized motion.
\subsection{Model-based Gravity Compensation Controller Design}

Gravity compensation controllers are seldom used in soft robot-based exoskeletons. In this work, we introduce such a controller designed to maintain the 2-DOF robot arm in its current position as dictated by the designers' motion settings (Fig. \ref{fig_12}(b)). The controller operates on the premise that both the center of mass and the mass of the human dummy are known. These parameters were inputted into the inverse quasi-static model to calculate the necessary force. This force, determined by the dummy arm's specifications, the real bending angle ($\theta\_real$), and the arm's current posture ($\theta\_Arm\_posture$), was used to compute the required pressure ($P\_desire$). During the experimental phase, human testers manipulated the joints of the 2-DOF human dummy arm into three distinct positions to evaluate the efficacy of this gravity compensation controller. The results of these tests are demonstrated in the accompanying video. 

In the experimental sections, the human testers moved the joints of the 2-DOF human dummy arm to three different positions to test the performance of this gravity compensation controller. The results are in the attached video.